\documentclass{article}
\PassOptionsToPackage{round,compress}{natbib}

\usepackage[preprint]{neurips_2026}
\makeatletter
\renewcommand{\@noticestring}{%
  Preprint. Project page: \url{https://wangjl-nb.github.io/AdaOcc_web/}%
}
\makeatother

\usepackage[utf8]{inputenc} 
\usepackage[T1]{fontenc}    
\usepackage{hyperref}       
\usepackage{url}            
\usepackage{booktabs}       
\usepackage{amsfonts}       
\usepackage{nicefrac}       
\usepackage{microtype}      
\usepackage{xcolor}         
\usepackage{graphicx}
\usepackage{multirow}
\usepackage{array}
\usepackage{wrapfig}
\usepackage{makecell}
\usepackage{enumitem}

\title{AdaOcc: Adaptive 3D Occupancy Prediction for Embodied Tasks}

\author{%
  Jinglong Wang\textsuperscript{1,2,$*$,$\S$} \quad
  Yunjie Wang\textsuperscript{2,3,$*$,$\S$} \quad
  Zhiyang Zhang\textsuperscript{1,2,$*$,$\S$} \\[0.35em]
  Jiawei He\textsuperscript{2,4,$\dagger$} \quad
  Ye Yuan\textsuperscript{5} \\[0.35em]
  Bo Qiu\textsuperscript{6,$\dagger$} \quad
  Jing Zhang\textsuperscript{1,$\dagger$} \\[0.8em]
  \textsuperscript{1}School of Software, Beihang University \quad
  \textsuperscript{2}Beijing Academy of Artificial Intelligence \\[0.25em]
  \textsuperscript{3}Hebei University of Technology \quad
  \textsuperscript{4}XYZ Embodied AI \\[0.25em]
  \textsuperscript{5}ShanghaiTech University \quad
  \textsuperscript{6}University of Science and Technology Beijing \\[0.6em]
  \texttt{wjlzy@buaa.edu.cn} \quad
  \texttt{202421902001@stu.hebut.edu.cn} \quad
  \texttt{20377279@buaa.edu.cn} \\[0.2em]
  \texttt{jwhe2024@gmail.com} \quad
  \texttt{yuanye2024@shanghaitech.edu.cn} \\[0.2em]
  \texttt{qiubo@ustb.edu.cn} \quad
  \texttt{zhang\_jing@buaa.edu.cn} \\[0.5em]
  \textsuperscript{$*$}These authors contributed equally to this work. \\[0.2em]
  \textsuperscript{$\S$}These authors conducted this work during an internship at XYZ Embodied AI. \\[0.2em]
  \textsuperscript{$\dagger$}Corresponding authors: Jing Zhang, Bo Qiu, and Jiawei He.
}

\begin{document}

\maketitle

\begin{abstract}

Embodied tasks demand accurate, flexible, and semantically rich 3D scene representations. 3D semantic occupancy is well suited to this requirement, as it can model holistic 3D spaces by encoding geometric occupancy along with semantic categories. However, existing occupancy prediction methods struggle to meet practical deployment requirements, such as adapting to varying computing budgets, sensor setups, and observation views. In this paper, we propose a point-based Adaptive 3D Occupancy Prediction method, called AdaOcc, tailored for embodied scenarios. To accommodate heterogeneous sensor inputs, AdaOcc uses an adaptive geometry-guided dual-branch encoder that can support RGB images in various numbers of views with (estimated) depth maps or LiDAR scans. AdaOcc represents occupied regions via sparse semantic points trained with a progressive query learning strategy, allowing the prediction computational budget to be flexibly adjusted through query point numbers and decoder layers. To facilitate high-fidelity geometric modeling for lightweight point-based occupancy learning, we further propose a novel containment loss that regularizes predicted points to reside within valid occupied regions. Extensive experiments show that our method achieves a new state-of-the-art on Occ-ScanNet with considerable performance improvements over previous methods. Moreover, our framework demonstrates strong practical applicability as an adaptive 3D perception module in real-world embodied systems.

\end{abstract}

\section{Introduction}
\begin{figure}[t]
    \centering
    \includegraphics[width=\linewidth]{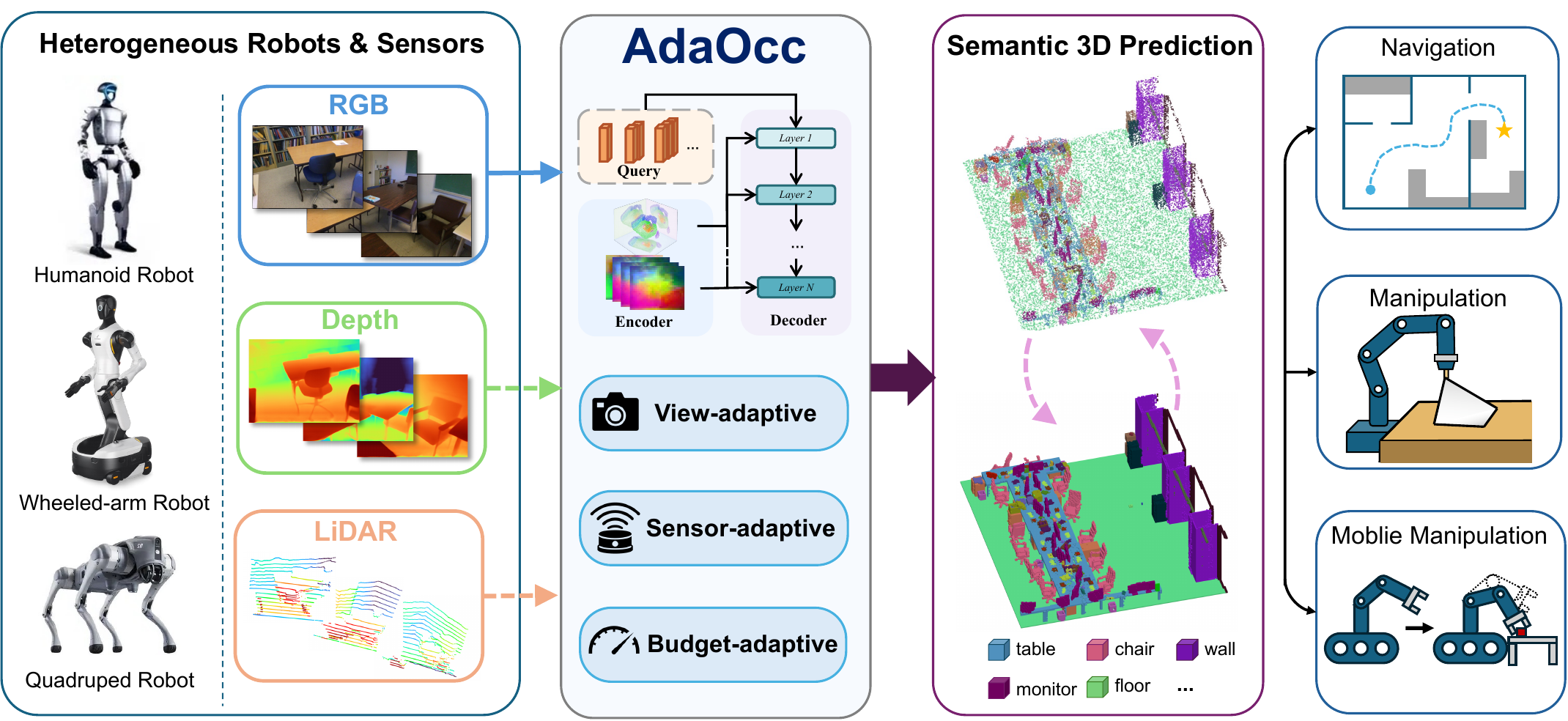}
    \vspace{-1em}
    \caption{
        Overview of AdaOcc. AdaOcc targets adaptive 3D occupancy prediction for embodied scene understanding. It accepts heterogeneous visual and geometric inputs from different embodied platforms and adapts to different computational budgets through adjustable query numbers and decoder depths. AdaOcc predicts sparse semantic points, which can be converted into semantic occupancy to provide structured spatial cues for downstream embodied tasks.
        }
    \label{fig:teaser}
\vspace{-2em}
\end{figure}

Accurate, flexible, and semantically rich 3D scene representations play a fundamental role in embodied AI tasks, as embodied agents typically make decisions based on their perception of the environment. 3D semantic occupancy can model holistic 3D spaces by encoding both geometric occupancy and semantic categories, providing a general-purpose representation for embodied scene understanding~\citep{song2017semantic,cao2022monoscene,li2023voxformer,zhang2023occformer,wei2023surroundocc,tong2023scene}. 

Early 3D occupancy prediction methods mainly focus on outdoor driving scenes, representing 3D scenes as dense voxel grids~\citep{zhang2023occformer,wei2023surroundocc,tong2023scene,huang2023tri,yu2023flashocc,ma2024cotr,lu2024octreeocc}, 3D Gaussian primitives~\citep{huang2024gaussianformer,huang2025gaussianformer}, or sparse points~\citep{wang2024opus}. Recent research on embodied perception~\citep{wang2024embodiedscan,wu2025embodiedocc,wang2025embodiedocc++,zhang2025roboocc} extends these outdoor methods by improving geometric and semantic modeling for more diverse and complex indoor scenes, with concurrent work further exploring adaptive serialization~\citep{wang2026adasformer}, geometry-prior sparse Gaussians~\citep{zhou2026gpose}, and open-vocabulary prediction~\citep{zhou2026legoocc}. Despite these advances, prevailing methods still face inherent limitations in balancing accuracy and latency. Moreover, they cannot adapt to robotic platforms with heterogeneous sensor setups, and show limited generalization across diverse embodied downstream tasks.

In this work, we introduce AdaOcc, a unified point-based framework for adaptive 3D occupancy prediction in embodied tasks. As illustrated in Fig.~\ref{fig:teaser}, AdaOcc supports different robots and sensors, and can be applied to different downstream embodied tasks.
Since robotic platforms can have different numbers of cameras and different geometric sensors, we design an adaptive geometry-guided dual-branch encoder that accepts RGB observations with varying numbers of views and incorporates geometric cues from estimated depth maps, depth-camera measurements, or LiDAR scans. 
We unify different geometric cues into a calibrated 3D point cloud, by calibrating the LiDAR point cloud to the RGB camera coordinate system and lifting depth maps into point clouds, enabling AdaOcc to adapt to different sensing conditions.

We observe that robotic platforms differ in computational capability, and embodied tasks demand different levels of perceptual granularity (e.g., manipulation often requires finer perception than navigation). We design AdaOcc to initialize sparse point queries from geometric inputs and reconstruct occupied semantic points via a multi-layer decoder equipped with progressive query learning. This design supports adaptive inference by adjusting the number of query points or using intermediate decoder outputs, allowing AdaOcc to balance accuracy and latency under varying computational budgets and environmental complexity.


To further enhance geometric fidelity in our point-based pipeline, we introduce a containment loss to regularize predicted points to reside within valid occupied regions, reducing surface-floating artifacts and improving spatial consistency around object boundaries and occupied structures. Experiments on Occ-ScanNet~\citep{yu2024monocular} show that AdaOcc achieves new state-of-the-art performance, reaching 65.29 IoU and 59.67 mIoU and outperforming the strongest prior method by 2.15 and 3.48 points, respectively. Further efficiency, adaptability, and real-world deployment studies demonstrate that AdaOcc is not only a strong benchmark model, but also a practical spatial perception module for downstream embodied tasks.
Our main contributions are summarized as follows:

\begin{itemize}[leftmargin=*]
    \item We propose AdaOcc, a point-based adaptive 3D occupancy framework for embodied scene understanding. 
    AdaOcc enables flexible occupancy prediction under varying compute budgets, sensor setups, and inference settings.
    
    \item For robust point-based occupancy prediction, we introduce key technical designs including an adaptive geometry-guided dual-branch encoder, progressive query learning for adaptive inference, and containment-guided optimization for improved boundary fidelity.
    
    \item AdaOcc achieves state-of-the-art performance on Occ-ScanNet with large margins over previous methods. Extensive experiments and real-world embodied-system deployment demonstrate its effectiveness, efficiency, and adaptability for downstream embodied tasks.
\end{itemize}

\vspace{-0.5em}
\section{Related work}
\vspace{-0.5em}

\paragraph{Semantic occupancy prediction.}

Early 3D scene understanding research is largely framed as semantic scene completion, which recovers complete scene geometry and semantics from partial RGB-D observations~\citep{song2017semantic,garbade2019two,li2019depth,roldao2020lmscnet,chen20203d,cai2021semantic}. This formulation later moves toward vision-only scene understanding, where monocular and camera-based methods infer 3D semantic occupancy from RGB observations by learning to lift image features into 3D space~\citep{cao2022monoscene,yao2023ndc,li2023voxformer,yu2024monocular,wang2026adasformer}. As occupancy prediction expands to larger scenes and multi-view settings, the field further shifts from dense voxel prediction toward more efficient representations. Dense voxel and view-transformation methods provide regular 3D supervision but suffer from high memory and computation costs~\citep{zhang2023occformer,wei2023surroundocc,tong2023scene,wang2024panoocc,li2023fb}, motivating compact decompositions, sparse structures, Gaussian primitives, and query-based formulations for scalable 3D reasoning~\citep{yu2023flashocc,ma2024cotr,lu2024octreeocc,tang2024sparseocc,huang2025gaussianformer,jia2023occupancydetr,li2024viewformer,wang2024opus,zhou2026gpose,zhou2026legoocc}. These works establish strong occupancy prediction backbones, but their assumptions on inputs, views, and inference budgets are usually tied to fixed evaluation protocols.

\vspace{-0.5em}
\paragraph{Occupancy for embodied perception.}
Recent embodied AI studies further extend occupancy prediction from offline scene reconstruction toward deployable spatial perception for robot navigation, interaction, and decision-making. Embodied perception benchmarks emphasize holistic 3D representations for realistic agent environments~\citep{wang2024embodiedscan,xu2024memory}, while robot-oriented occupancy methods begin to consider online perception, temporal scene understanding, and embodied semantic reasoning~\citep{wu2025embodiedocc,wang2025embodiedocc++,zhang2025roboocc,li2025sliceocc}. These works suggest that 3D occupancy is a useful spatial representation for embodied perception and robot scene understanding. However, practical robot deployment introduces additional variability that is less explored in existing studies. Most current embodied occupancy pipelines are still built around fixed sensing inputs, fixed view configurations, and fixed prediction budgets, making them difficult to adapt across platforms and tasks without redesign or retraining. Motivated by this gap, AdaOcc develops adaptive point-based occupancy prediction for embodied deployment.

\vspace{-0.5em}
\section{Method}
\label{sec:method}

\begin{figure*}[t]
    \centering
    \includegraphics[width=\textwidth]{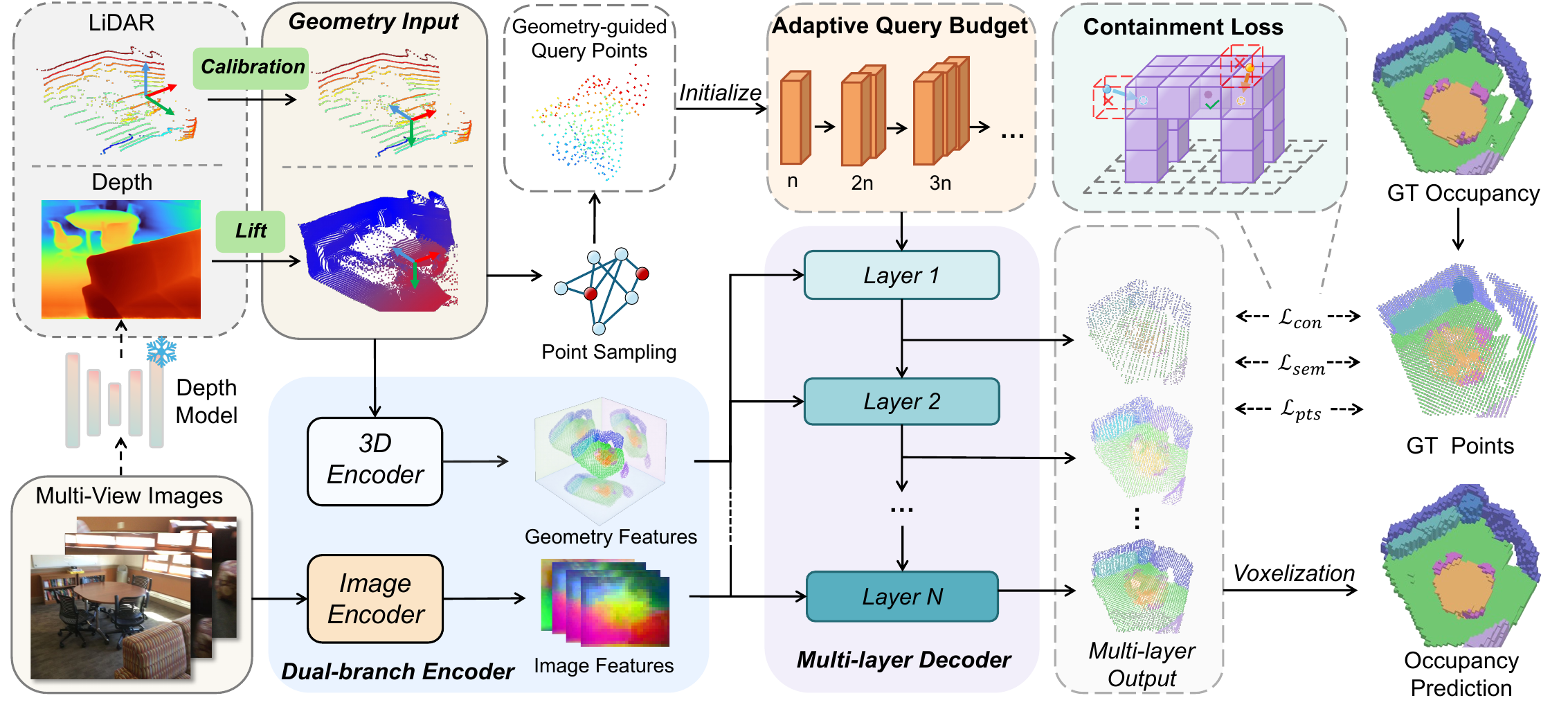}
    \caption{Overview of AdaOcc. AdaOcc is an adaptive 3D occupancy prediction framework for embodied scene understanding. It takes RGB observations as the primary input and can optionally incorporate geometric cues from estimated depth maps, depth cameras, or LiDAR scans. AdaOcc also adjusts its prediction budget under different computational constraints. The predicted semantic occupancy provides structured spatial cues for downstream embodied tasks.}
    \label{fig:pipeline}
    \vspace{-1em}
\end{figure*}

\subsection{Problem Setup and Method Overview}

We study semantic occupancy prediction in an agent-centric 3D space. Given one or multiple calibrated RGB images $\mathcal{I}=\{\mathbf{I}^{v}\}_{v=1}^{V}$ and optional geometric observations $\mathcal{G}=\{\mathbf{g}_{m}\}_{m=1}^{M}$, the goal is to predict the occupancy state and semantic label within a bounded local 3D region $\Omega$. Here, $\mathcal{G}$ is a unified point set in the local agent-centric coordinate system, where $M$ is the number of geometric points and $\mathbf{g}_{m}$ is the $m$-th point. When no independent geometric sensor is available, $\mathcal{G}$ can be estimated solely from $\mathcal{I}$, as in our Occ-ScanNet experiments. Following standard settings, $\Omega$ is discretized into a voxel grid $\mathcal{V}=\{\mathbf{p}_{i}\}_{i=1}^{H\times W\times D}$. Each voxel has a label $y_i\in\{0,1,\ldots,C\}$, where $0$ denotes empty space and $1,\ldots,C$ denote occupied semantic classes. Instead of directly classifying all voxels, AdaOcc predicts a sparse semantic point set $\hat{\mathcal{P}}=\{(\hat{\mathbf{p}}_{j},\hat{\mathbf{s}}_{j})\}_{j=1}^{N}$, where $\hat{\mathbf{p}}_{j}$ is a predicted occupied location and $\hat{\mathbf{s}}_{j}$ contains its semantic logits. Finally, the predicted point set is voxelized into $\hat{\mathbf{Y}}$ for standard occupancy evaluation.



As illustrated in Fig.~\ref{fig:pipeline}, AdaOcc is a lightweight point-based framework for adaptive 3D occupancy prediction in embodied scenes. It uses an adaptive geometry-guided dual-branch encoder to process RGB observations and optional geometric cues from estimated depth maps, depth cameras, or LiDAR scans. Geometry-guided queries are then refined by a progressive point decoder, producing sparse semantic points under adjustable query and decoder budgets. The model is trained with multi-layer point supervision and the proposed containment-guided optimization.

\subsection{Adaptive Geometry-Guided Dual-Branch Encoding}

Embodied occupancy prediction requires semantic cues from RGB observations and spatial grounding from geometric measurements. AdaOcc therefore adopts an adaptive geometry-guided dual-branch encoder that uses RGB observations as the primary input and optionally incorporates geometric observations when available. The two branches produce image and geometric features as:
\begin{equation}
\mathcal{F}^{\mathrm{img}}
=
E_{\mathrm{img}}(\mathcal{I})
=
\{\mathbf{F}^{\mathrm{img}}_{r}\}_{r=1}^{R_{\mathrm{img}}},
\qquad
\mathcal{F}^{\mathrm{geo}}
=
E_{\mathrm{geo}}(\mathcal{G})
=
\{\mathbf{F}^{xy}, \mathbf{F}^{xz}, \mathbf{F}^{yz}\}.
\end{equation}
Here, $\mathbf{F}^{\mathrm{img}}_{r}$ denotes the image feature map at the $r$-th scale, and $R_{\mathrm{img}}$ is the number of image feature scales. The geometric observations $\mathcal{G}$ are first voxelized and encoded as follows. For depth inputs, including estimated depth maps and depth-camera measurements, AdaOcc lifts depth pixels into 3D points using camera intrinsics and transforms them with the corresponding camera extrinsics. For LiDAR scans, AdaOcc transforms points from the LiDAR coordinate frame to the RGB-camera coordinate frame using calibrated sensor extrinsics, aligning geometric measurements with visual observations. After this normalization, different geometric sources share the same point-set interface and can be processed by the same geometric branch.

To encode geometry, AdaOcc voxelizes $\mathcal{G}$ and models its local 3D structure with a sparse 3D convolutional encoder. Instead of materializing a dense 3D feature volume, the resulting sparse geometric features are compressed into three orthogonal feature planes $\{\mathbf{F}^{xy},\mathbf{F}^{xz},\mathbf{F}^{yz}\}$. The $xy$ plane captures horizontal layout cues, while the $xz$ and $yz$ planes preserve vertical geometric profiles, providing height-aware spatial context with low memory cost.


\subsection{Progressive Point Query Decoding}
\begin{wrapfigure}{r}{0.55\textwidth}
    \vspace{-1em}
    \centering
    \includegraphics[width=0.5\textwidth]{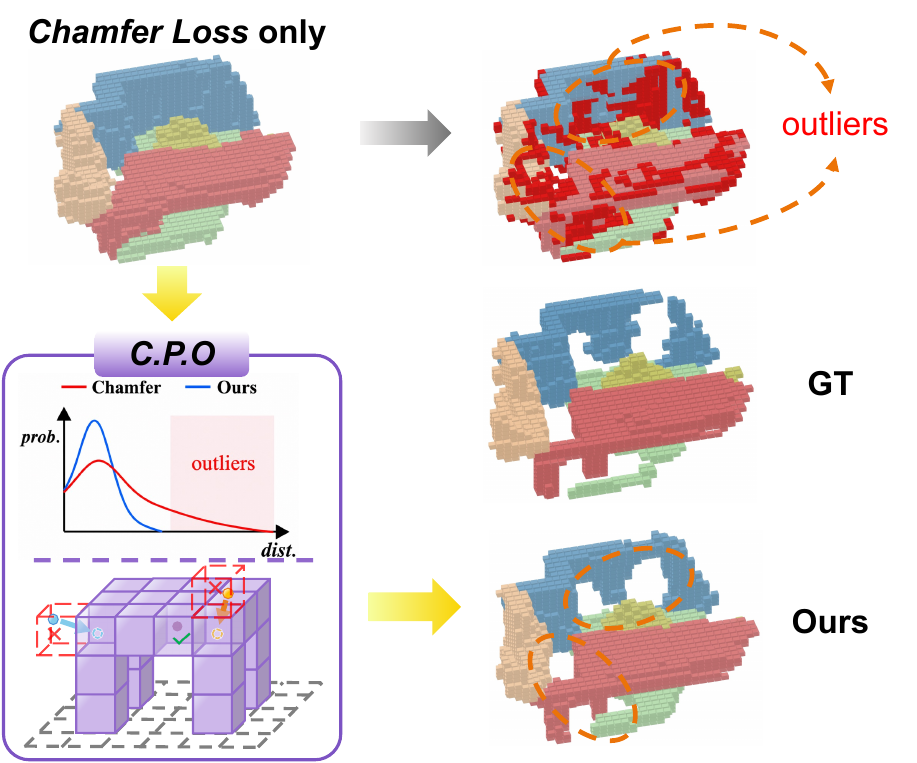}
    \vspace{-6pt}
    \caption{Illustration of Containment-Guided Point Optimization (C.P.O.), which regularizes predicted outlier points in supervised empty space to reside inside nearby occupied voxels, reducing surface-floating artifacts.}
    \label{fig:cpo}
    \vspace{-2em}
\end{wrapfigure}
Inspired by OPUS~\citep{wang2024opus}, AdaOcc adopts a sparse point decoder and extends it with geometry-guided query initialization and progressive query learning for budget-adaptive inference. Given an active query budget $N_q$, AdaOcc initializes point queries as $\mathcal{Q}_0=\{(\mathbf{q}^{0}_{i}, \mathbf{f}^{0}_{i})\}_{i=1}^{N_q}$ by sampling 3D locations from the geometric point set $\mathcal{G}$, where $\mathbf{q}^{0}_{i}\in\Omega$ is the initial reference location and $\mathbf{f}^{0}_{i}$ is the query feature. This provides sensor-agnostic spatial anchors because $\mathcal{G}$ can come from estimated depth maps, depth cameras, or LiDAR scans.

 At decoder layer $\ell$, each query predicts $K$ 3D sampling coordinates around its current reference location, denoted as $\mathcal{R}_{i}^{(\ell)}=\{\mathbf{r}_{i,k}^{(\ell)}\}_{k=1}^{K}$. AdaOcc then samples both appearance and geometric evidence at these coordinates: $\mathbf{z}^{\mathrm{img}}_{i,k}=\psi_{\mathrm{img}}(\mathbf{r}_{i,k}^{(\ell)},\mathcal{F}^{\mathrm{img}})$ and $\mathbf{z}^{\mathrm{geo}}_{i,k}=\psi_{\mathrm{geo}}(\mathbf{r}_{i,k}^{(\ell)},\mathcal{F}^{\mathrm{geo}})$. Here, $\psi_{\mathrm{img}}$ projects each sampling coordinate onto calibrated image planes and samples multi-scale image features, while $\psi_{\mathrm{geo}}$ samples geometric features from the $xy$, $xz$, and $yz$ planes.
The sampled appearance and geometric features are aggregated by the decoder and fused with the previous query feature through a residual update:
\begin{equation}
\mathbf{f}_{i}^{(\ell)}
=
\mathbf{f}_{i}^{(\ell-1)}
+
\phi_{\ell}
\left(
\mathbf{f}_{i}^{(\ell-1)},
\{(\mathbf{z}^{\mathrm{img}}_{i,k},\mathbf{z}^{\mathrm{geo}}_{i,k})\}_{k=1}^{K}
\right),
\end{equation}
where $\phi_{\ell}$ denotes the feature aggregation and mixing operation in the $\ell$-th decoder layer.

The updated query feature predicts $N_p^{(\ell)}$ semantic points at layer $\ell$, denoted as $\{(\hat{\mathbf{p}}^{(\ell)}_{i,j},\hat{\mathbf{s}}^{(\ell)}_{i,j})\}_{j=1}^{N_p^{(\ell)}}$, where $\hat{\mathbf{p}}^{(\ell)}_{i,j}=\mathbf{q}_{i}^{(\ell-1)}+\Delta\mathbf{p}^{(\ell)}_{i,j}$ is a predicted occupied location and $\hat{\mathbf{s}}^{(\ell)}_{i,j}$ denotes semantic logits. These predicted points serve as both the layer-wise occupancy output and the basis for spatial refinement. We recenter the next-layer query reference location by the mean predicted coordinate, $\mathbf{q}_{i}^{(\ell)}=\frac{1}{N_p^{(\ell)}}\sum_{j=1}^{N_p^{(\ell)}}\hat{\mathbf{p}}^{(\ell)}_{i,j}$. This iterative feature and coordinate refinement allows sparse queries to progressively align with occupied regions.

The active query number directly controls the prediction budget. To improve robustness across query budgets, AdaOcc gradually increases the active number of queries during training:
\begin{equation}
N_q(t)=
\min\left(
N_{\max},
N_{\mathrm{init}}
+
\left\lfloor
\frac{t-t_0}{\Delta t}
\right\rfloor
\Delta N
\right),
\end{equation}
where $t$ is the training epoch, $N_{\mathrm{init}}$ is the initial query budget, $\Delta N$ is the query increment, $\Delta t$ is the growth interval, and $N_{\max}$ is the maximum query budget. This curriculum enables the same trained decoder to adapt to different computational budgets by changing $N_q$ or using intermediate decoder outputs at inference time.

To handle varying observation views, we apply view-level masking during training by randomly retaining a subset of RGB views and their associated geometric observations. When $V_t$ views are retained, the total query budget becomes $N_q(t,V_t)=V_t N_q(t)$, allowing the model to process different numbers of views jointly in a single forward pass.

\subsection{Containment-Guided Point Optimization}

Point-based occupancy prediction is commonly supervised by Chamfer-style point-set reconstruction losses, such as Chamfer Distance~\citep{fan2017point} and Density-aware Chamfer Distance (DCD)~\citep{wu2021density}. Although these losses encourage predicted points to cover the target geometry, they do not explicitly enforce whether a point lies inside an occupied region. As shown in Fig.~\ref{fig:cpo}, this may produce surface-floating points that are close to occupied voxels but still located in empty space, leading to inaccurate free-space boundaries.

To address this issue, we introduce Containment-Guided Point Optimization (C.P.O.), which adds a containment loss to regularize predicted points to reside within valid occupied regions. Let $\mathcal{P}^{+}=\{\mathbf{p}_{i}\in\mathcal{V}\mid y_i>0\}$ denote occupied ground-truth voxel centers. Each occupied voxel center $\mathbf{p}_i$ induces an occupied box with side length $\mathbf{v}$, where $\mathbf{v}$ is the voxel size. For a predicted point $\hat{\mathbf{p}}$, its distance to the nearest occupied box is
\begin{equation}
d_{\mathrm{box}}(\hat{\mathbf{p}}, \mathcal{P}^{+})
=
\min_{\mathbf{p}_i\in\mathcal{P}^{+}}
\left\|
\mathrm{ReLU}
\left(
|\hat{\mathbf{p}}-\mathbf{p}_i|-\frac{\mathbf{v}}{2}
\right)
\right\|_2 ,
\end{equation}
where $|\cdot|$ and the subtraction by $\frac{\mathbf{v}}{2}$ are applied element-wise. The distance is zero if $\hat{\mathbf{p}}$ lies inside any occupied voxel box and increases as it drifts into empty space.

Let $\Omega_{\mathrm{occ}}\subseteq\Omega$ denote the valid occupancy supervision region defined by the evaluation mask. For decoder layer $\ell$, we collect predictions that fall into supervised empty space:
\begin{equation}
\mathcal{O}^{(\ell)}
=
\left\{
\hat{\mathbf{p}}^{(\ell)}
\in
\hat{\mathcal{P}}^{(\ell)}
\mid
\hat{\mathbf{p}}^{(\ell)} \in \Omega_{\mathrm{occ}},
\;
Y(\hat{\mathbf{p}}^{(\ell)})=0
\right\}.
\end{equation}
The containment loss is defined as
\begin{equation}
\mathcal{L}_{\mathrm{con}}
=
\frac{1}{|\mathcal{S}|}
\sum_{\ell\in\mathcal{S}}
\frac{1}{|\mathcal{O}^{(\ell)}|}
\sum_{\hat{\mathbf{p}}\in\mathcal{O}^{(\ell)}}
\rho_{\beta}
\left(
d_{\mathrm{box}}(\hat{\mathbf{p}}, \mathcal{P}^{+}) + m
\right),
\end{equation}
where $\mathcal{S}$ denotes the supervised decoder layers, $\rho_{\beta}$ is the Smooth-L1 penalty, and $m$ is a small positive offset that keeps the penalty active near occupied-box boundaries, encouraging predictions to move inside occupied voxels rather than stopping near the surface. By penalizing only predictions in supervised empty regions, the loss complements point-set reconstruction without suppressing valid occupied predictions or unsupervised space.

\subsection{Training Objective}

AdaOcc is trained with multi-layer supervision over semantic point predictions. At each decoder layer $\ell$, the predicted point set $\hat{\mathcal{P}}^{(\ell)}$ is supervised by semantic classification, point-set reconstruction, and containment regularization. The overall objective is
\begin{equation}
\mathcal{L}
=
\sum_{\ell=1}^{L}
\gamma^{L-\ell}
\left(
\lambda_{\mathrm{sem}}\mathcal{L}_{\mathrm{sem}}^{(\ell)}
+
\lambda_{\mathrm{pts}}\mathcal{L}_{\mathrm{pts}}^{(\ell)}
\right)
+
\lambda_{\mathrm{con}}\mathcal{L}_{\mathrm{con}},
\end{equation}
where $\mathcal{L}_{\mathrm{sem}}^{(\ell)}$ is focal loss~\citep{lin2017focal} over semantic logits, $\mathcal{L}_{\mathrm{pts}}^{(\ell)}$ is DCD point reconstruction between $\hat{\mathcal{P}}^{(\ell)}$ and the occupied ground-truth point set $\mathcal{P}^{+}$, and $\mathcal{L}_{\mathrm{con}}$ is the proposed containment loss. The containment term is applied to selected late decoder layers in C.P.O. $L$ denotes the number of decoder layers, $\gamma$ is the layer-wise decay factor, and $\lambda_{\mathrm{sem}}$, $\lambda_{\mathrm{pts}}$, and $\lambda_{\mathrm{con}}$ balance the semantic, reconstruction, and containment terms. 
\section{Experiments}
\label{sec:experiments}

\subsection{Datasets and Evaluation Metrics}


We evaluate AdaOcc on Occ-ScanNet, an indoor semantic occupancy prediction benchmark built upon RGB-D scans~\citep{dai2017scannet}. The task requires predicting a semantic occupancy grid for a bounded local 3D volume from calibrated RGB observations. AdaOcc predicts over an agent-centric range that is discretized into a dense voxel grid of size $130 \times 120 \times 140$ with a voxel size of $0.08\,\mathrm{m}$, which fully contains the official Occ-ScanNet evaluation grid of size $60 \times 60 \times 36$ at the same voxel size. Following the benchmark protocol, metrics are computed on this official grid, and predictions falling outside it are not evaluated. The semantic label space contains 11 occupied categories, including ceiling, floor, wall, window, chair, bed, sofa, table, tvs, furniture, and objects, together with an empty class; voxels with unknown ground-truth labels are ignored during evaluation.

We report results on both Occ-ScanNet and Occ-ScanNet-mini. The mini split is used for efficient comparison and ablation studies, while the full split is used to evaluate the final model against existing methods. Following the standard protocol, we use occupied IoU to measure binary occupancy quality and mean IoU (mIoU) over the 11 occupied semantic classes to evaluate semantic occupancy prediction. We also report per-class IoU to analyze category-level performance.



\subsection{Experimental Setup}

\renewcommand{\rothead}[1]{%
  \rotatebox[origin=c]{90}{\hspace{0.40em}\textbf{#1}\hspace{0.40em}}%
}
\newcolumntype{V}{!{\vrule width 0.5pt}}

\begin{table*}[!t]
\centering
\caption{Local prediction performance on the Occ-ScanNet dataset. \textbf{Rep.} denotes the main scene representation: $\mathcal{V}$ for dense voxels, $\mathcal{T}$ for TPV features, $\mathcal{G}$ for Gaussians, and $\mathcal{P}$ for points or point queries. ${}^{\dagger}$ indicates the AdaOcc variant with an EfficientNet image encoder. Bold and underline indicate the best and second-best results.}
\vspace{1em}
\small
\setlength{\tabcolsep}{4pt}
\renewcommand{\arraystretch}{1.08}
\resizebox{\textwidth}{!}{%
\begin{tabular}{cVcVcVcVcccccccccccVc}
\toprule
\textbf{Dataset} & \textbf{Method} & \textbf{Rep.} & \textbf{mIoU}
& \rothead{ceiling}
& \rothead{floor}
& \rothead{wall}
& \rothead{window}
& \rothead{chair}
& \rothead{bed}
& \rothead{sofa}
& \rothead{table}
& \rothead{tvs}
& \rothead{furniture}
& \rothead{objects}
& \textbf{IoU} \\
\noalign{\vskip 2pt}
\midrule
\noalign{\vskip 2pt}

\multirow{16}{*}{Occ-ScanNet}
& TPVFormer       & $\mathcal{T}$ & 24.94             & 6.96              & 32.97             & 14.41             & 9.10              & 24.01             & 41.49             & 45.44             & 28.61             & 10.66             & 35.37             & 25.31             & 33.39             \\
& MonoScene       & $\mathcal{V}$ & 24.62             & 15.17             & 44.71             & 22.41             & 12.55             & 26.11             & 27.03             & 35.91             & 28.32             & 6.57              & 32.16             & 19.84             & 41.60             \\
& ISO             & $\mathcal{V}$ & 28.71             & 19.88             & 41.88             & 22.37             & 16.98             & 29.09             & 42.43             & 42.00             & 29.60             & 10.62             & 36.36             & 24.61             & 42.16             \\
& SurroundOcc     & $\mathcal{V}$ & 30.83             & 18.90             & 49.30             & 24.80             & 18.00             & 26.80             & 42.00             & 44.10             & 32.90             & 18.60             & 36.80             & 26.90             & 42.52             \\
& GaussianFormer  & $\mathcal{G}$ & 29.93             & 20.70             & 42.00             & 23.40             & 17.40             & 27.00             & 44.30             & 44.80             & 32.70             & 15.30             & 36.70             & 25.00             & 40.91             \\
& OPUS            & $\mathcal{P}$ & 38.96             & 39.06             & 45.04             & 34.97             & 28.63             & 35.92             & 49.27             & 54.39             & 37.93             & 23.93             & 45.04             & 34.42             & 45.62             \\

& AdaSFormer      & $\mathcal{V}$ & 45.33 & 37.66 & 57.34 & 40.51 & 29.49 & 43.29 & 60.41 & 63.08 & 47.05 & 29.63 & 54.08 & 36.15 & 54.63 \\

& EmbodiedOcc     & $\mathcal{G}$ & 45.48             & 40.90             & 50.80             & 41.90             & 33.00             & 41.20             & 55.20             & 61.90             & 43.80             & 35.40             & 53.50             & 42.90             & 53.95             \\
& EmbodiedOcc++   & $\mathcal{G}$ & 46.20             & 36.40             & 53.10             & 41.80             & 34.40             & 42.90             & 57.30             & 64.10             & 45.20             & 34.80             & 54.20             & 44.10             & 54.90             \\
& DiScene         & $\mathcal{P}$ & 47.17             & 45.21             & 50.63             & 40.38             & 36.73             & 42.28             & 59.68             & 62.04             & 45.60             & 41.17             & 52.42             & 42.72             & 51.99             \\
& RoboOcc         & $\mathcal{G}$ & 47.67             & 45.36             & 53.49             & 44.35             & 34.81             & 43.38             & 56.93             & 63.35             & 46.35             & 36.12             & 55.48             & 44.78             & 56.48             \\

& SplatSSC        & $\mathcal{G}$ & 51.83 & 49.10 & \underline{59.00} & 48.30 & 38.80 & 47.40 & 62.40 & 67.00 & 49.50 & 42.60 & 60.70 & 45.40 & 62.83 \\

& GPOcc-DPT       & $\mathcal{G}$ & 51.88 & 51.42 & 50.35 & 46.97 & 41.84 & 46.98 & 60.39 & 66.16 & 50.51 & 47.97 & 58.88 & 49.23 & 56.96 \\
& GPOcc-VGGT      & $\mathcal{G}$ & 56.19 & 51.67 & \textbf{59.93} & 52.07 & 46.44 & 51.35 & 64.45 & 69.47 & 54.30 & 51.76 & 63.29 & 53.36 & 63.14 \\

& AdaOcc$^\dagger$ (Ours) & $\mathcal{P}$ & \underline{59.03}    & \underline{53.97}    & 57.90    & \underline{53.14}    & \underline{47.34}    & \underline{55.95}    & \underline{70.38}    & \underline{74.34}    & \underline{59.86}    & \underline{52.95} & \underline{66.90}    & \underline{56.66}    & \underline{64.60}    \\

& AdaOcc (Ours) & $\mathcal{P}$ & \textbf{59.67}    & \textbf{55.40}    & 58.50 & \textbf{53.63}    & \textbf{48.00}    & \textbf{56.81}    & \textbf{70.82}    & \textbf{74.83}    & \textbf{60.89}    & \textbf{53.15}    & \textbf{67.22}    & \textbf{57.17}    & \textbf{65.29}    \\
\noalign{\vskip 2pt}
\midrule
\noalign{\vskip 2pt}
\multirow{7}{*}{Occ-ScanNet-mini}
& MonoScene       & $\mathcal{V}$ & 25.90             & 17.00             & 46.20             & 23.90             & 12.70             & 27.00             & 29.10             & 34.80             & 29.10             & 9.70              & 34.50             & 20.40             & 41.90             \\
& ISO             & $\mathcal{V}$ & 29.40             & 21.10             & 42.70             & 24.60             & 15.10             & 30.80             & 41.00             & 43.30             & 32.20             & 12.10             & 35.90             & 25.10             & 42.90             \\
& EmbodiedOcc     & $\mathcal{G}$ & 45.57             & 29.50             & 49.40             & 41.70             & 36.30             & 41.90             & 60.40             & 59.60             & 46.30             & 34.50             & 58.00             & 43.50             & 55.13             \\
& EmbodiedOcc++   & $\mathcal{G}$ & 48.20             & 23.30             & 51.00             & 42.80             & 39.30             & 43.50             & 65.60 & 64.00 & 50.70 & 40.70    & 60.30             & 48.90 & 55.70             \\
& SplatSSC        & $\mathcal{G}$ & 48.87 & 36.60 & 55.70 & 46.50 & 40.10 & 45.60 & 64.50             & 62.40             & 48.60             & 30.60             & 61.20 & 45.39             & 61.47 \\
& AdaOcc$^\dagger$ (Ours) & $\mathcal{P}$ & \underline{57.97}    & \underline{45.76}    & \underline{57.46}    & \underline{56.15}    & \underline{47.50}    & \textbf{59.17}    & \underline{74.91}    & \textbf{75.16}    & \textbf{57.63}    & \underline{42.20} & \underline{64.18}    & \underline{57.54}    & \underline{65.16}    \\

& AdaOcc (Ours) & $\mathcal{P}$ & \textbf{58.74}    & \textbf{48.73}    & \textbf{57.61}    & \textbf{56.25}    & \textbf{48.18}    & \underline{58.90}    & \textbf{75.30}    & \underline{75.07}    & \underline{57.58}    & \textbf{46.52} & \textbf{64.37}    & \textbf{57.61}    & \textbf{65.38}    \\
\noalign{\vskip 2pt}
\bottomrule
\end{tabular}%
}
\label{tab:occ_scannet_local_prediction}
\end{table*}

\begin{figure*}[t]
    \centering
    \includegraphics[width=\textwidth]{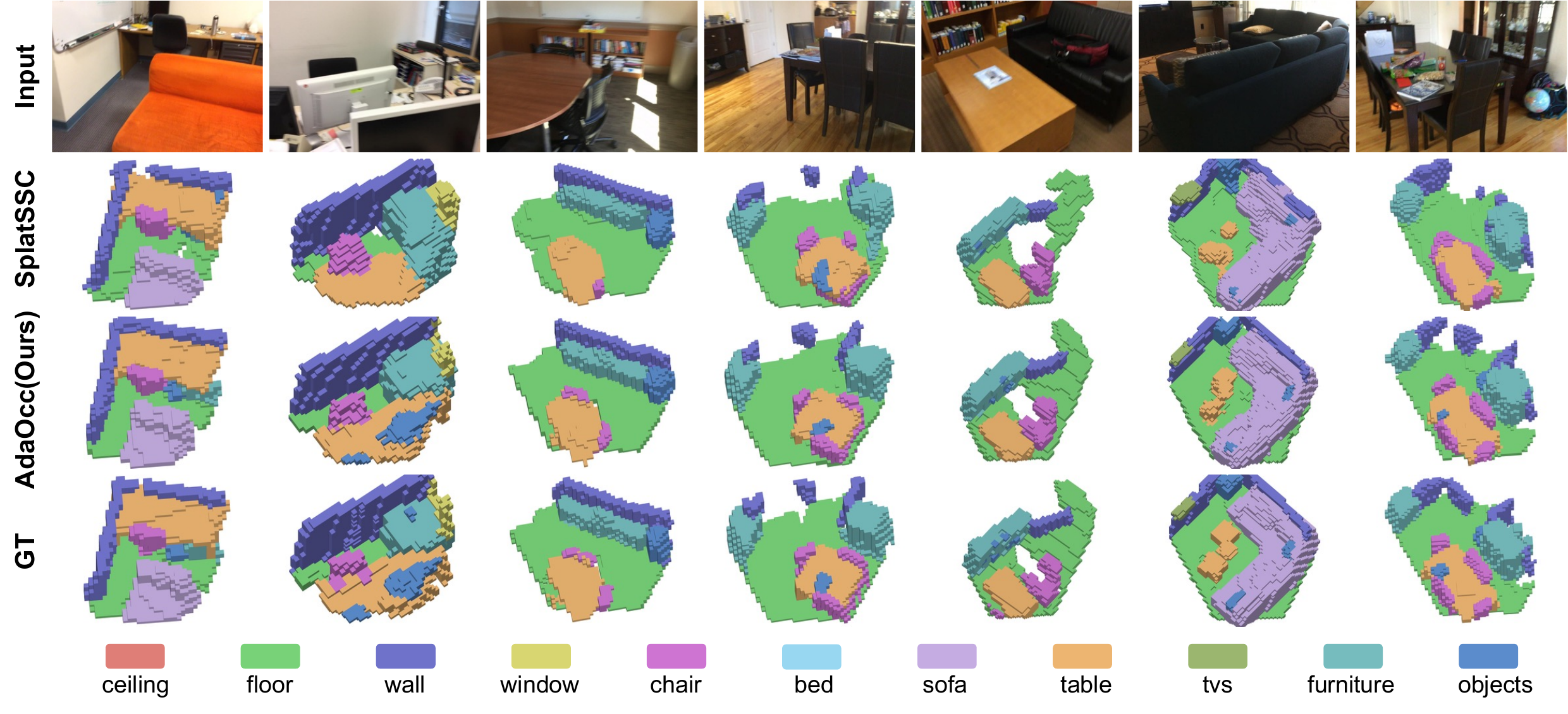}
    \caption{Qualitative comparison on Occ-ScanNet. AdaOcc produces cleaner and more complete semantic occupancy compared to the existing approaches.}
    \label{fig:qualitative_comparison}
\vspace{-1.8em} 
\end{figure*}

By default, AdaOcc uses C-RADIOv3-B~\citep{heinrich2025radiov2} as the image encoder and a sparse 3D convolutional encoder for geometric feature extraction. Following SplatSSC~\citep{qian2026splatssc}, our benchmark experiments use depth maps estimated from the input RGB observations by Depth Anything v2~\citep{yang2024depth}; the depth maps are back-projected into 3D points and used only for geometry-guided query initialization and geometric encoding. No ground-truth depth, point cloud, or occupancy information is used as input during inference. We quantify the effect of the depth estimator, including a second monocular estimator and ground-truth depth, in Appendix~\ref{sec:supp_depth}. Training follows the objective described in Sec.~\ref{sec:method}, and predicted points are voxelized for standard evaluation. AdaOcc$^{\dagger}$ denotes the variant that uses the same EfficientNet image encoder as SplatSSC~\citep{tan2019efficientnet,qian2026splatssc}, replacing the default C-RADIOv3-B encoder for a controlled comparison.

\subsection{Main Results}

Table~\ref{tab:occ_scannet_local_prediction} summarizes the main results on the Occ-ScanNet and Occ-ScanNet-mini benchmarks. On the full Occ-ScanNet split, AdaOcc achieves state-of-the-art performance with 59.67 mIoU and 65.29 IoU. The strongest prior method, GPOcc-VGGT~\citep{zhou2026gpose}, reaches 56.19 mIoU and 63.14 IoU, so AdaOcc improves over it by 3.48 mIoU and 2.15 IoU even though GPOcc-VGGT uses the stronger VGGT geometry~\citep{wang2025vggt} prior while AdaOcc estimates depth from RGB alone. For the controlled comparison we retain SplatSSC~\citep{qian2026splatssc} as the primary baseline because it shares our setting, estimating depth from RGB with Depth-Anything-V2, and is the strongest method under this prior: it reaches 62.83 IoU, whereas the Depth-Anything-V2 variant of GPOcc obtains 56.96, and its IoU is within 0.31 of GPOcc-VGGT despite the latter using the much stronger VGGT prior. Swapping the geometry prior inside GPOcc changes its own IoU by 6.18, indicating that the advantage of the strongest competitor is largely attributable to its prior. Relative to SplatSSC, AdaOcc improves by 7.84 mIoU and 2.46 IoU; the larger gain in mIoU than in IoU indicates stronger semantic discrimination across indoor categories. To control for the effect of the image backbone, AdaOcc$^{\dagger}$ uses the same EfficientNet image encoder as SplatSSC and still reaches 59.03 mIoU and 64.60 IoU, improving over SplatSSC by 7.20 mIoU and 1.77 IoU. This suggests that the gains mainly come from the progressive point decoder, the adaptive geometry-guided dual-branch encoder, and containment-guided point optimization, rather than from the image encoder. The qualitative results further support this observation: as shown in Fig.~\ref{fig:qualitative_comparison}, AdaOcc produces cleaner object-level structures and more complete indoor layouts.


\begin{figure*}[t]
    \centering
    \includegraphics[width=\textwidth]{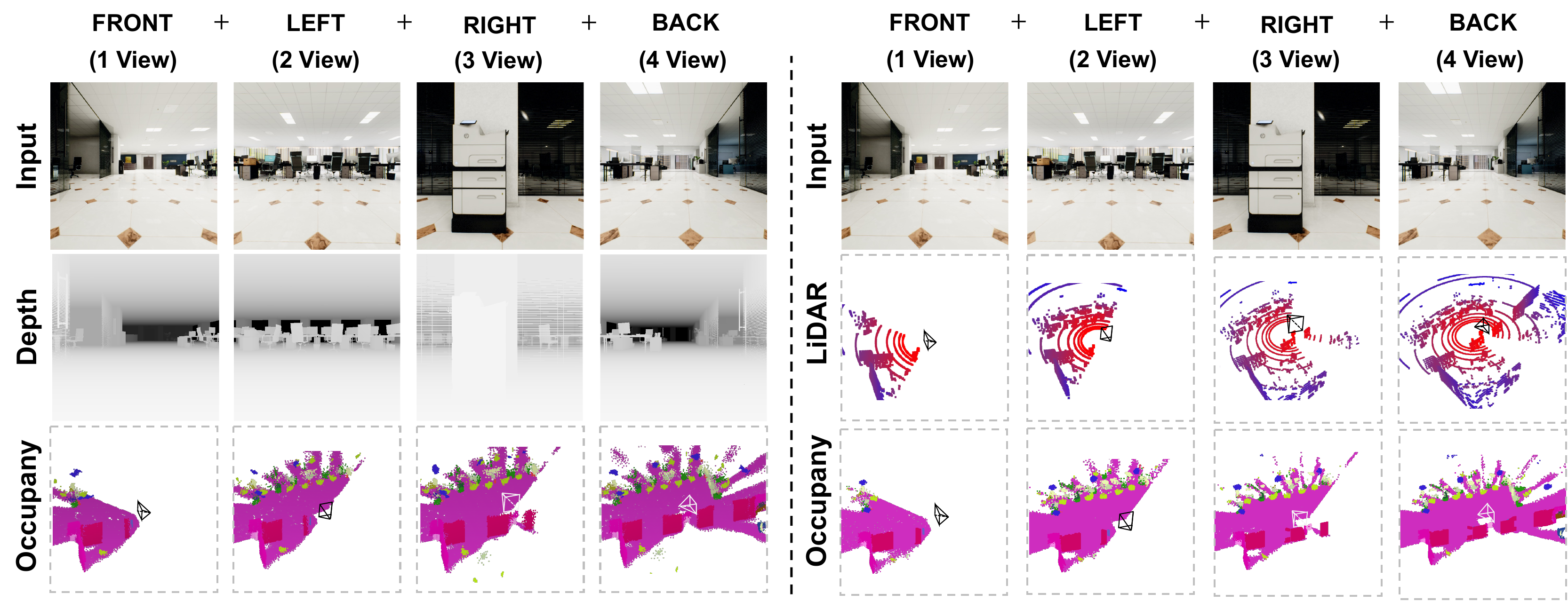}
    \vspace{-1.5em}
    \caption{Qualitative adaptability analysis on TartanGround. AdaOcc accepts geometric inputs from depth maps or LiDAR scans, and processes RGB observations with varying numbers of views jointly in a single forward pass.}
    \label{fig:tartanground_adaptability}
\vspace{-1em}
\end{figure*}

\subsection{Ablation Study}

\paragraph{Component Ablation}
\begin{wraptable}{hr}{0.55\textwidth}
\centering
\vspace{-2em}
\caption{Ablation study of key components in AdaOcc.}
\label{tab:ablation_components}
\small
\setlength{\tabcolsep}{3.5pt}
\renewcommand{\arraystretch}{1.08}
\resizebox{0.55\textwidth}{!}{%
\begin{tabular}{c!{\vrule width 0.5pt}cccc!{\vrule width 0.5pt}cc}
\toprule
\noalign{\vskip 2pt}
\makecell{\textbf{Image}\\\textbf{Encoder}} &
\makecell{\textbf{Geometric}\\\textbf{Initialization}} &
\makecell{\textbf{3D}\\\textbf{Encoder}} &
\makecell{\textbf{C.P.O.}} &
\makecell{\textbf{Progressive}\\\textbf{Training}} &
\makecell{\textbf{mIoU}} &
\makecell{\textbf{IoU}} \\
\noalign{\vskip 2pt}
\midrule
\noalign{\vskip 2pt}
\multirow{6}{*}{RADIO} & -- & -- & -- & -- & 49.65 & 56.29 \\
& -- & -- & $\checkmark$ & $\checkmark$ & 55.50 & 62.92 \\
& $\checkmark$ & -- & $\checkmark$ & $\checkmark$ & 57.06 & 64.72 \\
& $\checkmark$ & $\checkmark$ & -- & $\checkmark$ & 54.54 & 61.31 \\
& $\checkmark$ & $\checkmark$ & $\checkmark$ & -- & 54.63 & 62.35 \\
& $\checkmark$ & $\checkmark$ & $\checkmark$ & $\checkmark$ & \textbf{58.74} & \textbf{65.38} \\
\noalign{\vskip 2pt}
\midrule
\noalign{\vskip 2pt}
EffNet & $\checkmark$ & $\checkmark$ & $\checkmark$ & $\checkmark$ & \underline{57.97} & \underline{65.16} \\
\noalign{\vskip 2pt}
\bottomrule
\end{tabular}%
}
\vspace{-1em}
\end{wraptable}

Table~\ref{tab:ablation_components} ablates the key components of AdaOcc on Occ-ScanNet-mini. Starting from the RADIO image encoder baseline, adding C.P.O. and progressive training substantially improves performance from 49.65 mIoU and 56.29 IoU to 55.50 mIoU and 62.92 IoU. Adding geometric initialization further improves the result to 57.06 mIoU and 64.72 IoU, showing the benefit of sensor-guided spatial anchors. With the full 3D encoder, AdaOcc reaches 58.74 mIoU and 65.38 IoU, indicating that explicit geometric features provide complementary spatial grounding for point-based occupancy prediction. Removing either C.P.O. or progressive training causes clear performance drops, verifying the contribution of containment-guided optimization and budget-aware training. The EfficientNet variant remains competitive, confirming that the gains are not solely due to the image backbone.
\vspace{-6pt}
\paragraph{Hyperparameter sensitivity.}
Table~\ref{tab:hyper_sensitivity} varies the principal training hyperparameters on Occ-ScanNet-mini with all other settings fixed. The offset margin trades recall for precision: setting $m=0$ lowers IoU and mIoU by 1.13 and 1.27 points, while increasing it to $0.08$ changes them by only 0.06 and 0.62 points, so a small positive margin suffices. Halving or doubling the C.P.O. weight schedule changes IoU and mIoU by at most 0.19 and 0.58 points, and varying the decoder decay factor $\gamma$ between 0.80 and 0.95 changes them by at most 0.13 and 0.37 points. By contrast, removing progressive training costs 3.03 IoU and 4.11 mIoU, while using a finer query schedule changes the result by only 0.21 IoU and 0.59 mIoU. Overall, these results show that AdaOcc is robust to moderate variations of its loss-related hyperparameters.

\begin{table}[t]
\centering
\small
\caption{Sensitivity to training hyperparameters on Occ-ScanNet-mini. All other settings are fixed within each study, and bold marks the default configuration. For the query schedule, $X/Y$ denotes adding $X$ queries every $Y$ epochs, from $X$ up to 500.}
\label{tab:hyper_sensitivity}
\renewcommand{\arraystretch}{1.05}
\begin{tabular*}{0.8\linewidth}{@{\extracolsep{\fill}}lccc@{}}
\toprule
Parameter & Setting & IoU & mIoU \\
\midrule
Margin $m$ & 0.00 / \textbf{0.04} / 0.08 & 64.25 / \textbf{65.38} / 65.32 & 57.47 / \textbf{58.74} / 58.12 \\
C.P.O.\ weight & $0.5\times$ / \textbf{1$\times$} / $2\times$ & 65.19 / \textbf{65.38} / 65.32 & 58.16 / \textbf{58.74} / 58.17 \\
Decay $\gamma$ & 0.80 / \textbf{0.90} / 0.95 & 65.51 / \textbf{65.38} / 65.35 & 58.62 / \textbf{58.74} / 58.37 \\
Query schedule & -- / 50/20 / \textbf{100/40} & 62.35 / 65.17 / \textbf{65.38} & 54.63 / 58.15 / \textbf{58.74} \\
\bottomrule
\end{tabular*}
\end{table}

\subsection{Efficiency and Adaptability}

\begin{table}[t]
\centering
\caption{Efficiency and adaptability analysis on the Occ-ScanNet-mini dataset using a single NVIDIA H20 GPU. All results use the same EfficientNet image encoder as SplatSSC for a controlled comparison. (a) Query-number ablation for AdaOcc$^{\dagger}$. (b) Anchor-number ablation for SplatSSC. (c) Decoder-layer output ablation for AdaOcc$^{\dagger}$. Within each panel only the listed factor is varied while all other settings are fixed.}

\label{tab:ablation_efficiency}
\vspace{-0.6em}
\scriptsize
\setlength{\tabcolsep}{3pt}
\renewcommand{\arraystretch}{1.25}
\begin{minipage}[c]{0.31\textwidth}
\centering
\vspace{0.5em}
\textbf{(a) Query (Ours)}\\
\vspace{0.5em}
\resizebox{\linewidth}{!}{%
\begin{tabular}{@{}c@{\hspace{2pt}\vrule\hspace{2pt}}cl@{\hspace{2pt}\vrule\hspace{2pt}}cc@{}}
\toprule
\makecell{\textbf{Query}} &
\makecell{\textbf{Time$\downarrow$}\\\textbf{(ms)}} &
\makecell{\textbf{Mem.$\downarrow$}\\\textbf{(MiB)}} &
\makecell{\textbf{mIoU}} &
\makecell{\textbf{IoU}} \\
\midrule
200  & 63.2 & 1710.7 & 55.13 & 62.36 \\
\textbf{500}  & 64.4 & 1710.6 & 57.97 & \textbf{65.16} \\
800  & 73.3 & 1798.8 & \textbf{58.32} & 64.58 \\
1000 & 75.9 & 1863.7 & 58.25 & 64.29 \\
\bottomrule
\end{tabular}%
}
\end{minipage}%
\hfill
\begin{minipage}[c]{0.32\textwidth}
\centering
\vspace{0.5em}
\textbf{(b) Anchor (SplatSSC)}\\
\vspace{0.5em}
\resizebox{\linewidth}{!}{%
\begin{tabular}{@{}c@{\hspace{2pt}\vrule\hspace{2pt}}cl@{\hspace{2pt}\vrule\hspace{2pt}}cc@{}}
\toprule
\makecell{\textbf{Anchor}} &
\makecell{\textbf{Time$\downarrow$}\\\textbf{(ms)}} &
\makecell{\textbf{Mem.$\downarrow$}\\\textbf{(MiB)}} &
\makecell{\textbf{mIoU}} &
\makecell{\textbf{IoU}} \\
\midrule
600  & 91.0 & 1837.2 & 43.30 & 53.75 \\
\textbf{1131} & 85.1 & 1826.1 & \textbf{48.46} & \textbf{60.86} \\
1911 & 85.0 & 1835.1 & 45.26 & 58.85 \\
2596 & 85.6 & 1830.3 & 42.95 & 56.33 \\
\bottomrule
\end{tabular}%
}
\end{minipage}%
\hfill
\begin{minipage}[c]{0.31\textwidth}
\centering
\vspace{0.5em}
\textbf{(c) Decoder layers}\\ 
\vspace{0.5em}
\resizebox{\linewidth}{!}{%
\begin{tabular}{@{}c@{\hspace{2pt}\vrule\hspace{2pt}}cl@{\hspace{2pt}\vrule\hspace{2pt}}cc@{}}
\toprule
\makecell{\textbf{Layer}} &
\makecell{\textbf{Time$\downarrow$}\\\textbf{(ms)}} &
\makecell{\textbf{Mem.$\downarrow$}\\\textbf{(MiB)}} &
\makecell{\textbf{mIoU}} &
\makecell{\textbf{IoU}} \\
\midrule
3 & 58.2 & 1254.6 & 49.46 & 54.92 \\
4 & 59.9 & 1405.9 & 54.51 & 60.96 \\
5 & 62.5 & 1556.4 & 56.80 & 63.67 \\
\textbf{6} & 64.4 & 1710.6 & \textbf{57.97} & \textbf{65.16} \\
\bottomrule
\end{tabular}%
}
\end{minipage}
\vspace{-2em}
\end{table}

\begin{figure}[t]
    \vspace{-2em}
    \centering
    \includegraphics[width=\linewidth]{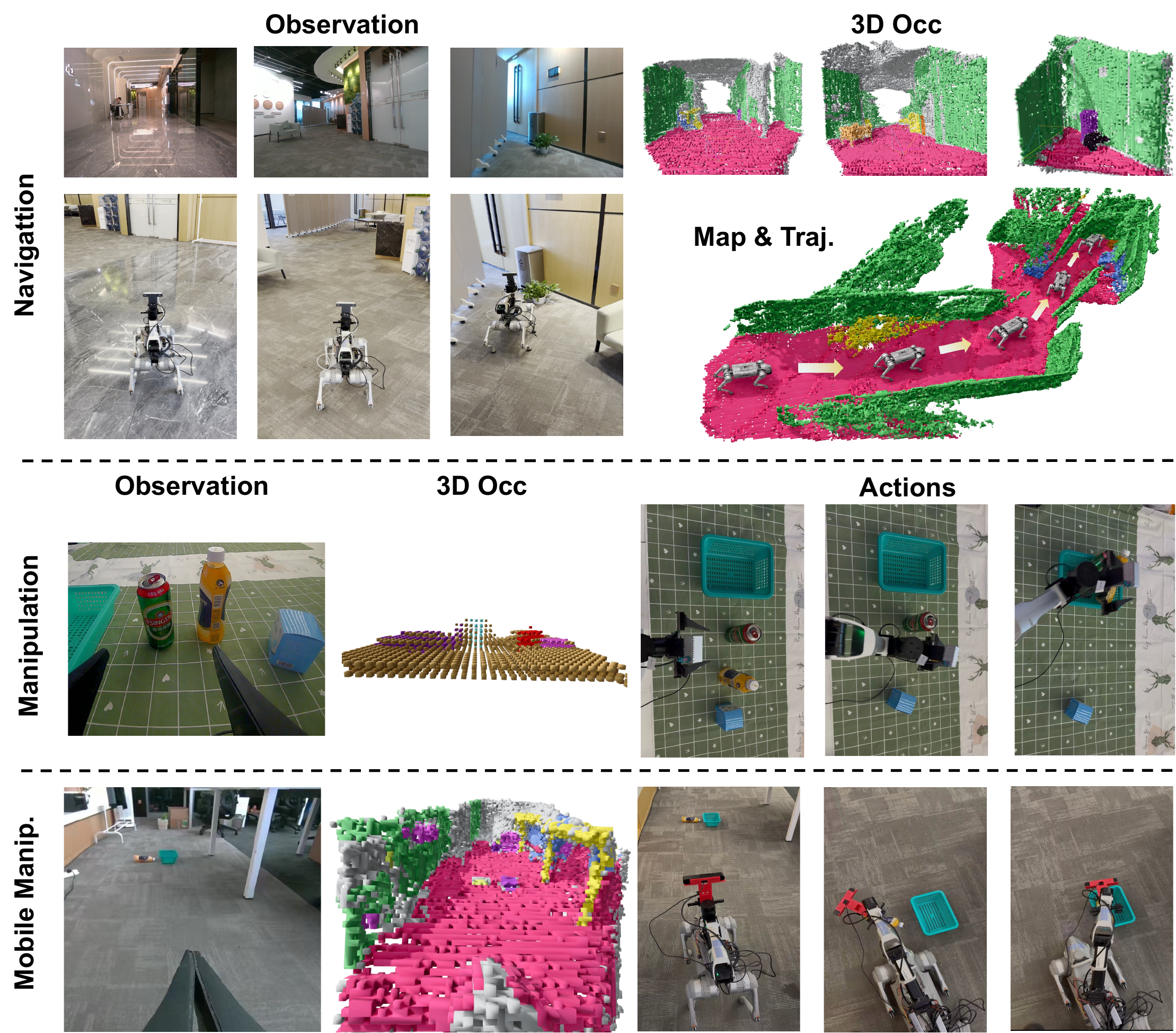}
    \caption{
    Real-world embodied applications of AdaOcc. AdaOcc provides semantic 3D occupancy for navigation, manipulation, and mobile manipulation. The semantic prompts are \textit{plant} for navigation, \textit{bottle} and \textit{basket} for manipulation, and \textit{trash} and \textit{basket} for mobile manipulation.
    }
    \label{fig:demo_application}
    \vspace{-1.5em}
\end{figure}

Table~\ref{tab:ablation_efficiency} evaluates the efficiency-adaptability trade-off on Occ-ScanNet-mini using a single NVIDIA H20 GPU. We use SplatSSC~\citep{qian2026splatssc} as the main efficiency baseline, since it reports lower latency and memory usage than EmbodiedOcc~\citep{wu2025embodiedocc} in its Occ-ScanNet-mini comparison. For a controlled comparison with SplatSSC, all results use the same EfficientNet image encoder, corresponding to AdaOcc$^{\dagger}$. AdaOcc$^{\dagger}$ is trained with a progressive query schedule from 100 to 500 queries, while SplatSSC is trained with 1131 anchors. When varying the query budget, AdaOcc remains stable across a wide range of query numbers and provides a stronger accuracy-latency trade-off than SplatSSC. For example, AdaOcc with 500 queries achieves 57.97 mIoU and 65.16 IoU at 64.4 ms, whereas the best SplatSSC setting obtains 48.46 mIoU and 60.86 IoU at 85.1 ms. The decoder-layer ablation further shows that the same trained model can produce valid intermediate predictions without retraining, enabling flexible latency-memory-accuracy trade-offs by selecting different decoder depths.

Beyond computational adaptability, Fig.~\ref{fig:tartanground_adaptability} shows that AdaOcc can handle heterogeneous geometric inputs and flexible multi-view observations on TartanGround~\citep{patel2025tartanground}. The same trained model produces coherent reconstructions with geometric inputs from either depth maps or LiDAR scans, and processes different numbers of RGB views jointly in a single forward pass rather than stitching per-view predictions.

\subsection{Embodied Applications}

\vspace{-0.5em}
To further examine the practical applicability of AdaOcc, we deploy it in real-world indoor embodied systems for navigation, manipulation, and mobile manipulation. As shown in Fig.~\ref{fig:demo_application}, AdaOcc reconstructs semantic 3D occupancy from robot observations and provides structured spatial cues for downstream planning and interaction. In navigation, the reconstructed occupancy supports target-aware path planning; in manipulation, it localizes task-relevant objects and receptacles; and in mobile manipulation, it supports navigation toward objects and subsequent placement. These demonstrations suggest that AdaOcc can serve as a practical 3D perception module beyond offline benchmark evaluation. More real-robot demonstrations are available at \url{https://wangjl-nb.github.io/AdaOcc_web/}, and a closed-loop navigation study with quantitative results is reported in Appendix~\ref{sec:supp_navigation}.

\vspace{-0.5em}
\section{Conclusions}
\vspace{-0.5em}
We presented AdaOcc, a unified point-based framework for adaptive 3D occupancy prediction in embodied scenes. AdaOcc normalizes geometric cues from estimated depth maps, depth-camera measurements, or LiDAR scans into a common point-set interface, and combines them with RGB observations from varying numbers of views through an adaptive geometry-guided dual-branch encoder. Its progressive point decoder supports budget-adaptive inference by changing query numbers or using intermediate decoder outputs, while the containment loss regularizes predicted points to reside within valid occupied regions and improves spatial consistency. Experiments on Occ-ScanNet and Occ-ScanNet-mini demonstrate state-of-the-art performance, strong efficiency-adaptability trade-offs, and practical deployment in a real robot navigation system. Future work will focus on improving the reconstruction of small and thin objects in cluttered embodied scenes.

\begin{ack}
\textbf{Funding.} This work was supported by the National Natural Science Foundation of China (No.~62461160331 and No.~62132001).

\textbf{Competing interests.} The first three authors conducted this work during an internship at XYZ Embodied AI. The authors declare no other competing financial interests or personal relationships that could have appeared to influence the work reported in this paper.
\end{ack}

\bibliographystyle{plainnat}   
\bibliography{references}

\newpage
\appendix
\section*{Supplemental Material}

\section{Additional Experiments}
\label{sec:supp_experiments}
\subsection{Open-Vocabulary Semantic Occupancy Prediction}

We further adapt AdaOcc to the open-vocabulary semantic occupancy setting. 
Different from conventional semantic occupancy prediction, which predicts a closed-set class label for each occupied voxel, our open-vocabulary occupancy prediction represents each occupied voxel with a text-aligned feature. 
This formulation enables arbitrary language queries over 3D occupied space, allowing semantically related
  concepts to activate consistent regions.
This makes the occupancy representation more flexible for downstream embodied tasks, where the queried object category may be expressed with synonyms or fine-grained natural-language descriptions.

To achieve this, we modify AdaOcc with a CLIP-aligned semantic feature branch. 
The original image backbone is replaced by a frozen OpenAI CLIP RN101 visual encoder~\cite{radford2021learning}. 
We directly connect the multi-stage ResNet features to an FPN~\cite{lin2017feature}.
The decoder is kept query-based, but each query predicts two outputs: a binary occupancy score and a semantic feature. 
The binary occupancy head models whether a query corresponds to occupied space, while the feature head maps occupied queries into the CLIP text embedding space.

For semantic supervision, we pre-compute a CLIP text prototype bank for the TartanGround vocabulary. 
For each class name $c$, we encode multiple prompt templates using the CLIP RN101 text encoder and average the normalized embeddings:
\[
\mathbf{t}_c =
\mathrm{Norm}\left(
\frac{1}{M}\sum_{m=1}^{M}
\mathrm{Norm}\left(E_{\mathrm{text}}(\pi_m(c))\right)
\right),
\]
where $\pi_m(\cdot)$ denotes a prompt template and $E_{\mathrm{text}}$ is the CLIP text encoder. 
During training, each occupied query matched to a ground-truth semantic voxel with label $y_i$ is supervised by the corresponding prototype $\mathbf{t}_{y_i}$.

The overall training loss is
\[
\mathcal{L}
=
\lambda_{\mathrm{occ}}\mathcal{L}_{\mathrm{occ}}
+
\lambda_{\mathrm{pts}}\mathcal{L}_{\mathrm{pts}}
+
\lambda_{\mathrm{con}}\mathcal{L}_{\mathrm{con}}
+
\lambda_{\mathrm{feat}}\mathcal{L}_{\mathrm{feat}} .
\]

The semantic feature loss is applied only to positive occupied queries:
\[
\mathcal{L}_{\mathrm{feat}}
=
\lambda_{\mathrm{cos}}
\left(1 -
\cos(\mathbf{f}_i, \mathbf{t}_{y_i})
\right)
+
\lambda_{\mathrm{ce}}
\mathrm{CE}
\left(
\frac{\mathbf{f}_i \mathbf{T}^{\top}}{\tau},
y_i
\right),
\]
where $\mathbf{f}_i$ is the predicted voxel feature, $\mathbf{T}$ is the CLIP prototype bank, and $\tau$ is the temperature. 
We use cosine alignment as the main supervision and a lightweight prototype classification term to stabilize the embedding space. 
Class-balanced weights are used to reduce the dominance of frequent classes, while rare categories are assigned stronger weights according to voxel statistics.

We train this open-vocabulary variant on the TartanGround indoor occupancy dataset~\cite{patel2025tartanground}. 
The voxel size is $0.08$m, and the model is trained for 200 epochs with the CLIP RN101 image encoder frozen. 
Only the FPN, occupancy decoder, binary occupancy head, and semantic feature head are optimized. 
At inference time, an arbitrary text query $q$ is encoded by the same CLIP text encoder, and each occupied voxel is scored by cosine similarity:
\[
s_i(q) =
\cos
\left(
\mathrm{Norm}(\mathbf{f}_i),
\mathrm{Norm}(E_{\mathrm{text}}(q))
\right).
\]

\begin{figure*}[t]
    \centering
    \includegraphics[width=0.8\textwidth]{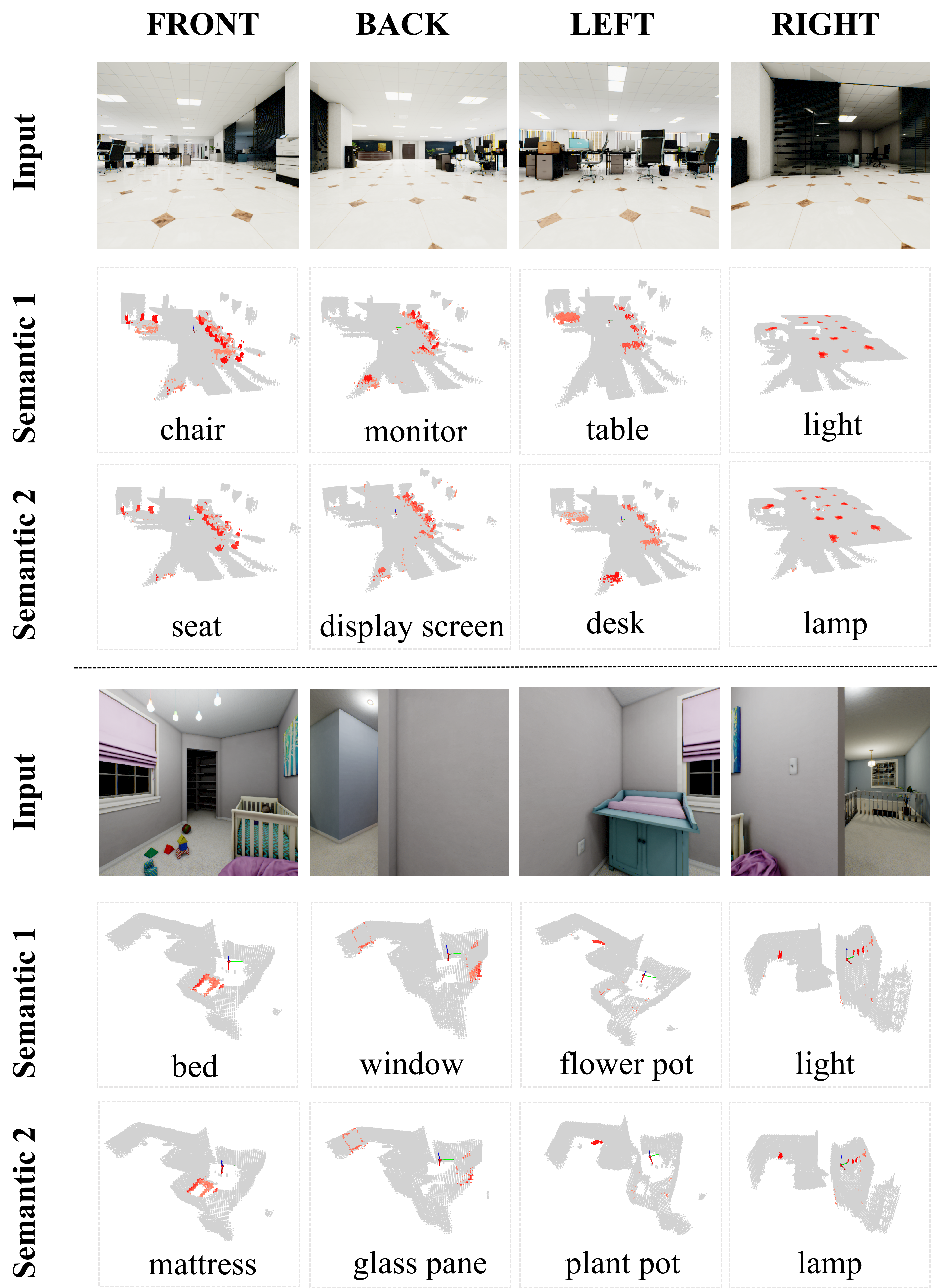}
    \caption{
    Open-vocabulary semantic activation in reconstructed 3D occupancy. 
    Semantically related text queries activate consistent object regions, demonstrating that the learned CLIP-aligned voxel features can respond to free-form language.
    }
    \label{fig:open_vocab_vis}
    \vspace{-1.5em}
\end{figure*}

Qualitatively, the trained model shows clear open-vocabulary behavior in 3D occupancy prediction. 
Given different but semantically related text queries, the CLIP-aligned voxel features tend to activate the same object regions in the reconstructed scene. 
For example, synonyms of objects such as beds and mattress produce consistent high-response areas. 
We visualize this by highlighting the top activated occupied points in red and rendering the remaining points in gray, as shown in Fig.~\ref{fig:open_vocab_vis}.

\begin{table}[t]
  \centering
  \caption{Comparison of the GT-depth and LiDAR input variants on the TartanGround test
  set.}
  \label{tab:pseudodepth_lidar_iou}
  \begin{tabular}{lccc}
  \toprule
  Variant & Input Source & IoU (\%) & mIoU (\%) \\
  \midrule
  GT-depth version & Depth-derived points & 67.55 & 26.13 \\
  LiDAR version &  LiDAR points & 55.65 & 19.40 \\
  \bottomrule
  \end{tabular}
\end{table}

\subsection{Flexible Multi-View and Geometric Input}
During open-vocabulary occupancy training, we further introduce a flexible input setting that supports varying numbers of camera views and geometric points from different sources, as illustrated in Fig.~\ref{fig:tartanground_adaptability}.
Although this is used in our open-vocabulary experiments, the same training strategy is general and can be applied to a broader range of experimental settings.

To enable this flexibility, we use four cameras, including the front, back, left, and right views. 
During training, we randomly select a subset of the available camera views as input, with the number of selected views ranging from one to four.
The model therefore sees different view combinations during training and can be evaluated with one, two, three, or four input views at inference time.

The geometric input can be provided either by depth-derived  points or by LiDAR points. 
For the RGB-D setting,  points are generated from the depth maps of the selected camera views. 
For the RGB-LiDAR setting, the raw LiDAR point cloud is used as the geometric input instead. 
When only a subset of camera views is selected, the same LiDAR point cloud is paired with the selected RGB views, while the image features are computed only from those views. 
The point loading and deduplication pipeline is adjusted accordingly, but the occupancy decoder and prediction heads remain unchanged.

With this design, the same training framework supports randomly sampled one-to-four-view input under both RGB-D and RGB-LiDAR settings. 
The model can thus reconstruct occupancy from different numbers of camera views while using either depth-based  points or LiDAR points as geometric support. We compare two settings on TartanGround test set. The results are shown in Table~\ref{tab:pseudodepth_lidar_iou}.

Table~\ref{tab:supp_views} reports the occupancy quality under one to four input views on 1{,}379 TartanGround test samples from 12 scenes (ground-truth depth). The model is trained once with randomly sampled view subsets and evaluated without retraining. Within the same observed regions, the reconstruction quality changes only slightly as the number of input views varies, indicating that AdaOcc is robust to different view configurations. Additional views mainly expand spatial coverage: under the fixed four-view union, the IoU increases from 17.24 with a single view to 67.55 with four views.

\begin{table}[h]
\centering
\small
\caption{Occupancy IoU / mIoU on the TartanGround test set under one to four input views, using the model trained with randomly sampled view subsets and evaluated without retraining. ``Current-view union'' evaluates the union of the regions covered by the views currently provided as input, whereas ``Four-view union'' uses the fixed union covered by all four cameras.}
\label{tab:supp_views}
\setlength{\tabcolsep}{4pt}
\renewcommand{\arraystretch}{1.05}
\resizebox{\linewidth}{!}{%
\begin{tabular}{lcccccc}
\toprule
Input views & Front & Left & Back & Right & Current-view union & Four-view union \\
\midrule
Front & 62.87 / 23.24 & -- & -- & -- & 62.87 / 23.24 & 17.24 / 6.93 \\
Front + Left & 65.98 / 25.56 & 64.43 / 23.95 & -- & -- & 65.29 / 25.06 & 32.34 / 13.29 \\
Front + Left + Back & 65.88 / 25.15 & 64.99 / 24.37 & 67.24 / 25.26 & -- & 66.14 / 25.04 & 52.14 / 19.18 \\
Four views & 67.57 / 26.55 & 66.33 / 25.18 & 68.63 / 26.49 & 67.29 / 27.03 & 67.55 / 26.14 & 67.55 / 26.14 \\
\bottomrule
\end{tabular}%
}
\end{table}

\subsection{Occ3D-nuScenes Dataset Results}
We conduct outdoor 3D semantic occupancy prediction experiments on Occ3D-nuScenes dataset. Compared with other SOTA methods, we achieve 3.8 RayIoU improvement. With LiDAR and surrounding RGB inputs, we have batter performance with 50.9 RayIoU, showing that we can utilize more geometry information and support different sensor and view inputs.
\begin{table}[t]
\centering
\caption{Comparison of RayIoU results on the Occ3D-nuScenes dataset.}
\label{tab:rayiou_comparison}
\vspace{0.5em}
\setlength{\tabcolsep}{4pt}
\renewcommand{\arraystretch}{1.15}
\begin{tabular}{lcccc}
\toprule
\textbf{Methods}  & \textbf{RayIoU$_{1m}$} & \textbf{RayIoU$_{2m}$} & \textbf{RayIoU$_{4m}$} & \textbf{RayIoU} \\
\midrule
RenderOcc~\citep{pan2024renderocc}         & 13.1 & 19.6 & 25.5 & 19.5 \\
BEVFormer~\citep{li2024bevformer}         & 26.1 & 32.9 & 38.0 & 32.4 \\
BEVDet-Occ         & 23.6 & 30.0 & 35.1 & 29.6 \\
BEVDet-Occ (8f)  & 26.6 & 33.1 & 38.2 & 32.6 \\
FB-Occ (16f)~\citep{li2023fb}     & 26.7 & 34.1 & 39.7 & 33.5 \\
SparseOcc (8f)~\citep{tang2024sparseocc}    & 28.0 & 34.7 & 39.4 & 34.0 \\
SparseOcc (16f)~\citep{tang2024sparseocc}  & 29.1 & 35.8 & 40.3 & 35.1 \\
\midrule
OPUS-T (8f)~\citep{wang2024opus}           & 31.7 & 39.2 & 44.3 & 38.4 \\
OPUS-S (8f)~\citep{wang2024opus}            & 32.6 & 39.9 & 44.7 & 39.1 \\
OPUS-M (8f)~\citep{wang2024opus}            & 33.7 & 41.1 & 46.0 & 40.3 \\
OPUS-L (8f)~\citep{wang2024opus}           & 34.7 & 42.1 & 46.7 & 41.2 \\
\midrule
AdaOcc (Ours) (1f) (RGB+Pseudo depth map)     & 37.55 & 46.14 & 51.20 & 44.97 \\
AdaOcc (Ours) (1f) (RGB+LiDAR)    & \textbf{46.60} & \textbf{51.58} & \textbf{54.48} & \textbf{50.89} \\
\bottomrule
\end{tabular}
\end{table}

\subsection{Hyperparameter Sensitivity}
\label{sec:supp_hyper}

We study the sensitivity of AdaOcc to its main training hyperparameters on Occ-ScanNet-mini. All variants keep the remaining settings fixed and are trained for 200 epochs.

\paragraph{Margin and C.P.O.\ weight.}
Table~\ref{tab:supp_margin} reports the effect of the C.P.O.\ offset margin $m$ and of the C.P.O.\ loss weight schedule. A positive margin converts a moderate recall reduction ($87.88\to83.41$) into a clear precision gain ($70.50\to75.32$), and $m=0.04$ attains the best IoU, mIoU, and F1 (79.16, against 78.24 for $m=0$ and 79.02 for $m=0.08$). Halving or doubling the weight schedule changes IoU and mIoU by at most 0.19 and 0.58 points, respectively.

\begin{table}[h]
\centering
\small
\caption{Effect of the C.P.O.\ offset margin $m$ and of the C.P.O.\ loss weight schedule on Occ-ScanNet-mini. Bold marks the default configuration.}
\label{tab:supp_margin}
\setlength{\tabcolsep}{5pt}
\renewcommand{\arraystretch}{1.05}
\begin{tabular}{llccccc}
\toprule
Study & Setting & IoU & mIoU & Precision & Recall & F1 \\
\midrule
\multirow{3}{*}{Margin $m$} & 0.00 & 64.25 & 57.47 & 70.50 & 87.88 & 78.24 \\
 & \textbf{0.04} & \textbf{65.38} & \textbf{58.74} & 75.32 & 83.41 & \textbf{79.16} \\
 & 0.08 & 65.32 & 58.12 & \textbf{75.36} & 83.06 & 79.02 \\
\midrule
\multirow{3}{*}{C.P.O.\ weight} & $0.05\to0.15$ & 65.19 & 58.16 & 74.32 & \textbf{84.14} & -- \\
 & $\textbf{0.10}\to\textbf{0.30}$ & \textbf{65.38} & \textbf{58.74} & \textbf{75.32} & 83.41 & -- \\
 & $0.20\to0.60$ & 65.32 & 58.17 & 75.17 & 83.29 & -- \\
\bottomrule
\end{tabular}
\end{table}

\paragraph{Decay factor and query schedule.}
Table~\ref{tab:supp_schedule} varies the decoder loss decay factor $\gamma$ and the granularity of the progressive query schedule. Both affect the final result only mildly: within $\gamma\in[0.80,0.95]$ the variation is at most 0.13 IoU and 0.37 mIoU, and using a finer query schedule instead of the default changes the result by only 0.21 IoU and 0.59 mIoU. Removing progressive training, by contrast, costs 3.03 IoU and 4.11 mIoU, indicating that the prediction budget, rather than the loss coefficients, is the dominant factor.

\begin{table}[h]
\centering
\small
\caption{Effect of the decoder loss decay factor $\gamma$ and of the progressive query schedule on Occ-ScanNet-mini. The query schedule is written as queries added per epoch interval; bold marks the default configuration.}
\label{tab:supp_schedule}
\setlength{\tabcolsep}{5pt}
\renewcommand{\arraystretch}{1.05}
\begin{tabular}{llcc}
\toprule
Study & Setting & IoU & mIoU \\
\midrule
\multirow{3}{*}{Decay $\gamma$} & 0.80 & 65.51 & 58.62 \\
 & \textbf{0.90} & \textbf{65.38} & \textbf{58.74} \\
 & 0.95 & 65.35 & 58.37 \\
\midrule
\multirow{3}{*}{Query schedule} & none & 62.35 & 54.63 \\
 & 50/20 & 65.17 & 58.15 \\
 & \textbf{100/40} & \textbf{65.38} & \textbf{58.74} \\
\bottomrule
\end{tabular}
\end{table}

\subsection{Depth Estimator Sensitivity}
\label{sec:supp_depth}

AdaOcc estimates depth from RGB with Depth Anything v2 (DAv2) by default. To quantify the dependence on this external component, we replace DAv2 with a second monocular estimator, MoGe-2, and, separately, replace the estimated depth with ground-truth depth, keeping the architecture, the evaluation protocol, and all training settings unchanged. We first measure depth quality over all 6{,}646 unique Occ-ScanNet-mini frames: both estimators process every frame, and predictions are resized to the ground-truth resolution and evaluated over valid depths.

\begin{table}[h]
\centering
\small
\caption{Depth quality and the resulting occupancy accuracy on Occ-ScanNet-mini. All three entries are produced in a single controlled run with identical training settings.}
\label{tab:supp_depth}
\setlength{\tabcolsep}{5pt}
\renewcommand{\arraystretch}{1.05}
\begin{tabular}{lcccc}
\toprule
Depth input & AbsRel $\downarrow$ & RMSE (cm) $\downarrow$ & IoU $\uparrow$ & mIoU $\uparrow$ \\
\midrule
Ground truth & 0.0000 & 0.00 & 69.61 & 62.67 \\
DAv2 & 0.0441 & 13.82 & 65.50 & 58.35 \\
MoGe-2 & 0.0782 & 23.43 & 62.44 & 55.76 \\
\bottomrule
\end{tabular}
\end{table}

Occupancy accuracy follows depth quality consistently: replacing DAv2 with the less accurate MoGe-2 reduces IoU and mIoU by 3.06 and 2.59 points, whereas ground-truth depth improves them by a further 4.11 and 4.32 points. The remaining headroom therefore lies in the external depth estimator, whose errors concentrate around reflective or low-texture surfaces, thin structures, and occluded boundaries. The DAv2 entry of Table~\ref{tab:supp_depth} stems from the run shared with the MoGe-2 and ground-truth variants and differs from the DAv2 results of Tables~\ref{tab:ablation_components} and~\ref{tab:hyper_sensitivity} (65.38 IoU / 58.74 mIoU) by training stochasticity.

\section{Embodied System}
\label{sec:supp_embodied_system}

\begin{figure*}[t]
    \centering
    \includegraphics[width=\textwidth]{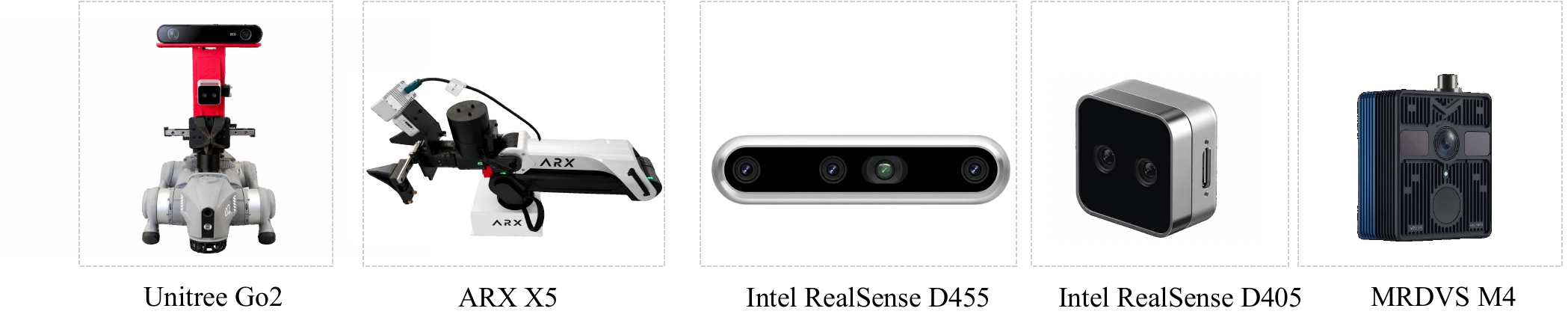}
    \caption{The photo of devices used in real-world embodied tasks.}
    \label{fig:device}
    \vspace{-1.5em}
\end{figure*}

We use different robots and perception sensors to conduct real-world embodied tasks, as listed in Table~\ref{tab:experimental_equipment}. The devices used in these tasks are shown in Fig.~\ref{fig:device}. The details of each task are as following: 
\subsection{Object Navigation with Go2}
We implement a lightweight embodied navigation system to connect AdaOcc predictions with real-robot execution. The platform uses a Unitree Go2 quadruped robot equipped with a D455 camera. AdaOcc reconstructs the observed scene as a semantic PLY point cloud, where each point is associated with its 3D position and semantic label or text-aligned semantic response. This semantic point cloud serves as the spatial representation for downstream navigation.

Given a language-specified target, we use open-vocabulary semantic activation in the reconstructed point cloud to determine the navigation goal. The text query activates semantically relevant occupied regions, and the goal position is obtained from the corresponding activated target region. In this way, the navigation module specifies goals through semantic concepts rather than manually defined metric coordinates.

To connect 3D reconstruction with robot navigation, we convert the semantic PLY map into a 2.5D traversability costmap. Floor or otherwise traversable points are projected onto a ground-plane grid as free space, while non-floor points within the robot collision-height range are treated as obstacles. The obstacles are further inflated according to the robot radius and a safety margin, yielding a conservative costmap for quadruped navigation. Unknown cells are treated as non-traversable during planning.

\begin{figure}[t]
    \centering
    \includegraphics[width=0.5\linewidth]{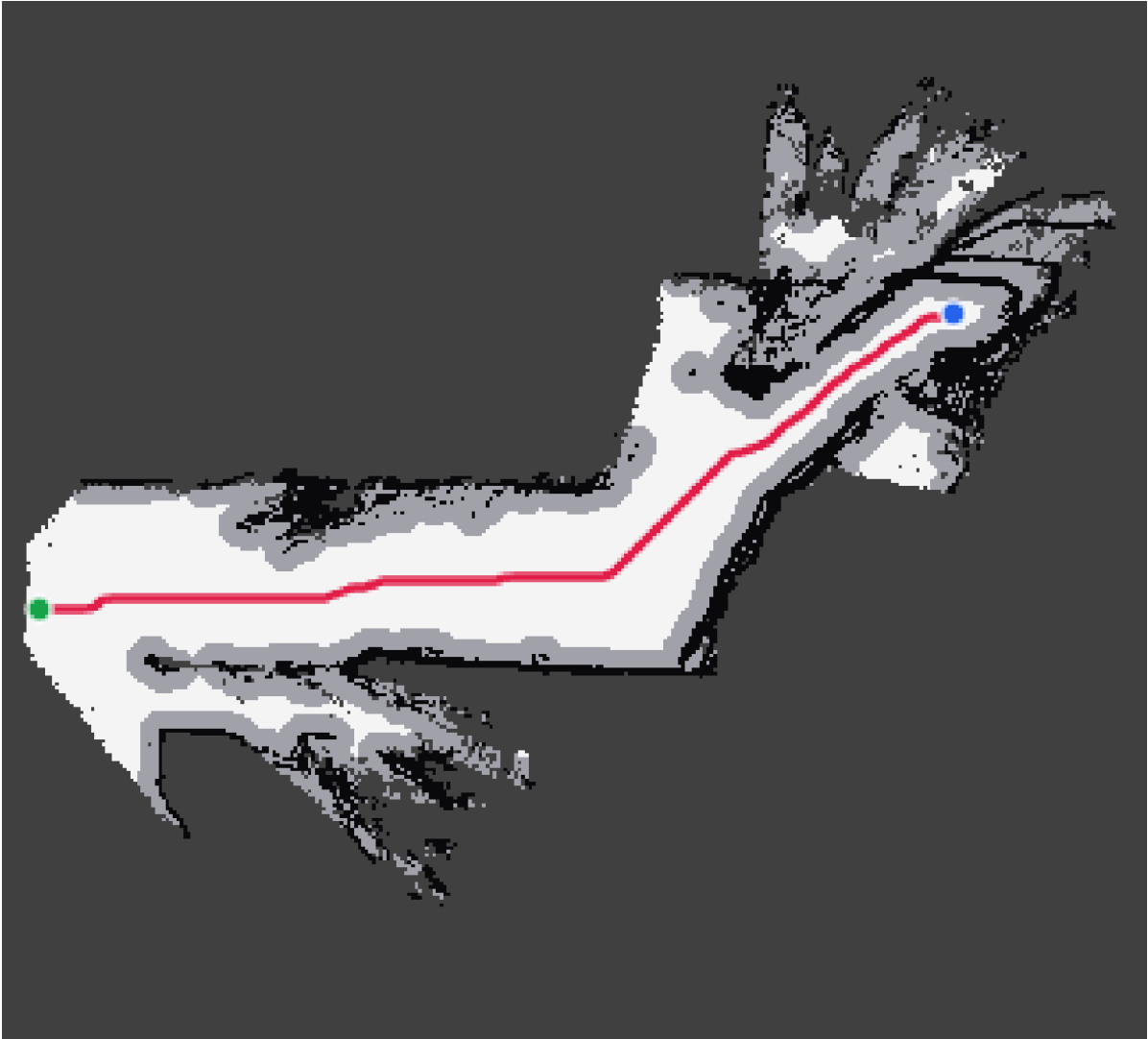}
    \caption{
    Costmap-based navigation from AdaOcc predictions. The semantic PLY prediction is converted into a 2.5D traversability costmap, where free space, inflated obstacles, and unknown regions are represented in different gray levels. Given an open-semantic target activation, A* plans a collision-aware path from the robot start position to the target region and exports map-frame waypoints for the Unitree Go2 robot.
    }
    \label{fig:supp_astar_plan}
\end{figure}

On the resulting costmap, we run A* search~\cite{hart1968formal} from the robot start position to the semantic target. The planner snaps the start and goal positions to nearby traversable cells when necessary, uses clearance-aware costs to prefer safer regions, and resamples the planned path into robot-friendly waypoints. The final output is a waypoint sequence in the map frame, including position, heading, speed, and accumulated path distance, which can be directly used by the Go2 waypoint tracker. Fig.~\ref{fig:supp_astar_plan} visualizes the costmap and the planned trajectory.

\subsection{Closed-Loop Navigation on FloorPlan-R2R}
\label{sec:supp_navigation}

We evaluate AdaOcc as the occupancy module of a closed-loop navigation policy on the FloorPlan-R2R val-unseen split in Habitat. Unlike standard vision-and-language navigation, the agent receives a semantic floorplan, in which each room is represented by a 2D polygon with a semantic category and a region ID, together with a concise instruction that specifies a start region, a target region, and an object- or relation-level stopping condition (e.g., ``start in office 16, go to living room 13, and stop at the couch''). At every step the agent observes egocentric RGB-D and selects among MOVE\_FORWARD, TURN\_LEFT, TURN\_RIGHT, and STOP, so success requires both reaching the correct region and stopping where the local observation satisfies the language condition. The agent is not given its pose in floorplan coordinates or the floorplan-to-world scale.

AdaOcc is used only as the online occupancy module: it converts accumulated egocentric RGB-D observations into a spatial representation used for floorplan grounding, target-region projection, and collision-free planning. The grounding, planning, and fixed zero-shot value-map stopping components are identical for both methods below, so the comparison isolates the utility of the occupancy representation.

\begin{table}[h]
\centering
\small
\caption{Closed-loop navigation on the FloorPlan-R2R val-unseen split. Both methods share the same grounding, planning, and stopping backend; FP-Nav additionally uses privileged floorplan alignment, whereas our pipeline estimates pose and scale online.}
\label{tab:supp_navigation}
\setlength{\tabcolsep}{5pt}
\renewcommand{\arraystretch}{1.05}
\begin{tabular}{lcccc}
\toprule
Method & NE $\downarrow$ & OSR $\uparrow$ & SR $\uparrow$ & SPL $\uparrow$ \\
\midrule
FP-Nav & 9.40 & 39.80 & 28.80 & 24.00 \\
AdaOcc + value-map stopping & \textbf{5.38} & \textbf{63.66} & \textbf{43.53} & \textbf{25.04} \\
\bottomrule
\end{tabular}
\end{table}

NE is the final distance to the goal, OSR records whether the agent reaches the success region at any time, SR additionally requires a correct final STOP, and SPL weighs SR by path efficiency. The AdaOcc occupancy representation reduces NE by 4.02\,m and improves OSR, SR, and SPL by 23.86, 14.73, and 1.04 points. The large OSR gain indicates more reliable cross-region navigation and the SR gain shows that this advantage persists through final task completion, while the smaller SPL gain reflects that stopping accuracy, rather than path efficiency, remains the main bottleneck.

This setting differs from occupancy prediction for autonomous driving, where benchmarks typically use a fixed, calibrated surround-view rig, predict a predefined metric volume around a vehicle in structured outdoor scenes, and evaluate occupancy mainly as a perception output. Here a moving embodied agent incrementally maps an unseen indoor environment from changing egocentric views, and narrow passages, nearby clutter, dense occlusion, and fine object or room boundaries directly affect action selection and stopping. The predicted occupancy therefore acts as persistent spatial memory and traversability evidence inside a closed-loop policy rather than as an endpoint in itself.

\subsection{Pick and Place with ARX X5}
We further validate the practical utility of AdaOcc in a real-world manipulation setup for picking bottle and place into basket using the ARX X5 platform. Given an input scene, AdaOcc predicts three object-level point sets, including two point clouds corresponding to the two target bottles and one point cloud corresponding to the basket.

The predicted point cloud of each bottle is independently provided to GraspNet~\cite{fang2020graspnet}, which outputs an executable grasp pose for the ARX X5 manipulator. This process yields the grasping poses for the two bottles shown in the figure. For placement, we compute the geometric center of the predicted basket point cloud and use it as the target placement position.

Based on the predicted bottle points, the generated grasp poses, and the basket-centered placement coordinate, the ARX X5 platform executes a complete manipulation pipeline for waste-bottle collection, including target grasping and basket placement. This experiment shows that the object-level geometric outputs of AdaOcc can be directly coupled with off-the-shelf grasp planning for real-robot manipulation. 

\subsection{Mobile Manipulation with ARX X5 on Go2}
Building on the navigation and manipulation pipeline above, we further demonstrate a long-range embodied mobile manipulation task using ARX X5 on Go2 platform. In this setting, the robot first observes a target bottle and a basket from a distance. The bottle location predicted by AdaOcc is used as the semantic navigation target, allowing Go2 to move from the initial observation point to the vicinity of the target object.

After reaching a suitable manipulation range, we apply the same object-centric perception and manipulation logic as above. Specifically, AdaOcc predicts the point sets of the target bottle and the basket in the local scene. The predicted bottle point set is then fed into GraspNet to generate an executable grasp pose, while the center of the predicted basket point set is used as the placement target. The robot then completes the bottle collection task by grasping the bottle and placing it into the basket.

This experiment demonstrates that AdaOcc supports a complete embodied pipeline that starts from long-range semantic target localization, continues with waypoint-based quadruped navigation, and finally enables goal-conditioned bottle grasping and basket placement in a real-world setting.

\begin{table*}[t]
\centering
\caption{Equipment Used in Our Real-world Experiments}
\vspace{0.8em}
\begin{tabular}{ccc}
\toprule
\textbf{Device Name} & \textbf{Type} & \textbf{Embodied Tasks} \\
\midrule
Unitree Go2 & Quadruped robot &
Navigation/Mobile manipulation \\
ARX X5 & Robotic arm &
Maniulation/Mobile manipulation  \\
Intel RealSense D455 & Long-range stereo depth camera &
Navigation\\
Intel RealSense D405 & Short-range depth camera &
Mobile manipulation \\
MRDVS M4 &  ToF RGB-D sensor &
Maniulation \\
\bottomrule
\end{tabular}%
\label{tab:experimental_equipment}
\end{table*}

\section{Standard Deviation of Experiments}
\label{sec:supp_deviation}
We repeat the best AdaOcc configuration on Occ-ScanNet-mini with five random seeds. S0--S4 correspond to seeds 0, 1785784487, 1602478221, 1383549716, and 977375669, respectively. As shown in Table~\ref{tab:supp_std_occscannet_mini}, AdaOcc shows stable performance across runs, with a standard deviation of 0.20 mIoU and 0.10 IoU. Across the five runs, the mIoU ranges from 58.26 to 58.74 and the IoU ranges from 65.24 to 65.49.

\begin{table*}[t]
\centering
\caption{Standard deviation over five random seeds on Occ-ScanNet-mini. S0--S4 correspond to seeds 0, 1785784487, 1602478221, 1383549716, and 977375669, respectively. All results are reported at epoch 200. Bold and underline indicate the best and second-best results among the five runs.}
\vspace{1em}
\small
\setlength{\tabcolsep}{4pt}
\renewcommand{\arraystretch}{1.08}
\resizebox{\textwidth}{!}{%
\begin{tabular}{cVcVcccccccccccVc}
\toprule
\textbf{Seed} & \textbf{mIoU}
& \rothead{ceiling}
& \rothead{floor}
& \rothead{wall}
& \rothead{window}
& \rothead{chair}
& \rothead{bed}
& \rothead{sofa}
& \rothead{table}
& \rothead{tvs}
& \rothead{furniture}
& \rothead{objects}
& \textbf{IoU} \\
\noalign{\vskip 2pt}
\midrule
\noalign{\vskip 2pt}
S0 & \textbf{58.74} & \textbf{48.73} & \textbf{57.61} & 56.25 & \textbf{48.18} & 58.90 & 75.30 & 75.07 & 57.58 & \textbf{46.52} & 64.37 & 57.61 & 65.38 \\
S1 & 58.26 & 47.13 & 57.39 & 56.12 & 47.64 & 58.69 & 75.19 & 74.99 & 57.63 & 43.60 & 64.47 & \textbf{58.05} & 65.24 \\
S2 & \underline{58.54} & \underline{48.69} & 57.54 & \underline{56.40} & \underline{48.15} & \underline{58.91} & \underline{75.53} & \underline{75.27} & \textbf{57.93} & 42.91 & 64.82 & 57.81 & \textbf{65.49} \\
S3 & 58.46 & 48.03 & 57.57 & 56.17 & 47.94 & \textbf{58.94} & \textbf{75.54} & 75.03 & \underline{57.85} & 43.18 & \textbf{65.19} & 57.59 & \underline{65.46} \\
S4 & 58.29 & 46.71 & \underline{57.59} & \textbf{56.51} & 47.52 & 58.88 & 74.71 & \textbf{75.28} & 57.30 & \underline{43.93} & \underline{64.92} & \underline{57.85} & \underline{65.46} \\
\noalign{\vskip 2pt}
\midrule
\noalign{\vskip 2pt}
Mean & 58.46 & 47.86 & 57.54 & 56.29 & 47.89 & 58.86 & 75.25 & 75.13 & 57.66 & 44.03 & 64.75 & 57.78 & 65.41 \\
Std. & 0.20 & 0.91 & 0.09 & 0.16 & 0.30 & 0.10 & 0.34 & 0.14 & 0.25 & 1.45 & 0.34 & 0.19 & 0.10 \\
\bottomrule
\end{tabular}%
}
\label{tab:supp_std_occscannet_mini}
\end{table*}

\section{Implementation Details and Compute Resources}
\label{sec:supp_resources}
\subsection{Implementation Details}
We provide the main implementation details for the Occ-ScanNet-mini AdaOcc configuration in Table~\ref{tab:supp_impl_details}. The model is trained for 200 epochs with progressive query learning, where the active query budget increases from 100 to 500 queries in five stages. During evaluation, the full query budget is used unless otherwise specified.

\paragraph{Coordinate frames.}
Each Occ-ScanNet sample provides camera intrinsics, camera-to-world extrinsics, ego-to-world pose, and LiDAR-to-ego pose. We use the ego/occupancy frame as the canonical local 3D frame for training. In the data loader, the original metadata contains the transform from ego to LiDAR, denoted as $T_{\mathrm{ego}\rightarrow\mathrm{lidar}}$, and an identity transform from ego to occupancy, $T_{\mathrm{ego}\rightarrow\mathrm{occ}}=I$. Depth-derived points are first reconstructed in the camera frame, transformed to the LiDAR frame using the corresponding sensor-to-LiDAR extrinsic, and then transformed into the occupancy frame by
\[
T_{\mathrm{lidar}\rightarrow\mathrm{occ}}
=
T_{\mathrm{ego}\rightarrow\mathrm{occ}}
T_{\mathrm{ego}\rightarrow\mathrm{lidar}}^{-1}.
\]
Since $T_{\mathrm{ego}\rightarrow\mathrm{occ}}$ is identity in this setting, the occupancy frame is aligned with the current ego frame. For image feature sampling, we construct an ego-to-image projection matrix $T_{\mathrm{ego}\rightarrow\mathrm{img}} = K T_{\mathrm{ego}\rightarrow\mathrm{cam}}$, where $K$ is the camera intrinsic matrix. Any image-space resize/crop transform is represented by a $4\times4$ image data augmentation matrix $A_{\mathrm{ida}}$ and left-multiplied into the projection matrix, i.e.,
\[
T_{\mathrm{ego}\rightarrow\mathrm{img}} \leftarrow A_{\mathrm{ida}} T_{\mathrm{ego}\rightarrow\mathrm{img}},
\]
so that the projected 3D points remain consistent with the resized input image.

\paragraph{Depth back-projection and pseudo points.}
For Occ-ScanNet dataset, we use the front camera only. The input RGB image has raw resolution $1296\times968$ and is resized/cropped to $960\times720$. The associated depth map is used to construct pseudo points. For each valid depth pixel $(u,v)$ with depth $d$, where $d\in[0.1,7.5]$ meters, we back-project using the OpenCV camera convention:
\[
x = \frac{(u+0.5-c_x)d}{f_x}, \qquad
y = \frac{(v+0.5-c_y)d}{f_y}, \qquad
z = d .
\]
The resulting camera-frame point $(x,y,z)$ is transformed to the LiDAR frame and then to the occupancy frame as described above. Each point stores five channels:
\[
(x,y,z,\mathrm{intensity},\Delta t),
\]
where the intensity channel is set to zero and $\Delta t$ is the timestamp offset. In the reported single-frame configuration, only the current frame is kept, so $\Delta t=0$ for the selected points. Points outside the local occupancy range are removed before voxelization.

\paragraph{Occupancy range and voxelization.}
The local 3D range is
\[
[-3.2,-4.8,-5.6,\,7.2,4.8,5.6],
\]
corresponding to $(x_{\min},y_{\min},z_{\min},x_{\max},y_{\max},z_{\max})$ in meters. We use a voxel size of
\[
\Delta = (0.08,0.08,0.08)\ \mathrm{m}.
\]
Thus the Cartesian grid has size
\[
N_x=130,\quad N_y=120,\quad N_z=140.
\]
This range is the agent-centric volume over which AdaOcc predicts; it fully contains the official Occ-ScanNet evaluation grid of size $60 \times 60 \times 36$ at the same voxel size.

Input points are voxelized with voxel size $0.08$ m, maximum 10 points per voxel, and at most 90k/120k voxels for train/test respectively. The voxel coordinates are computed by
\[
\mathbf{i}
=
\left\lfloor
\frac{\mathbf{p}_{\mathrm{occ}}-\mathbf{p}_{\min}}{\Delta}
\right\rfloor ,
\]
where $\mathbf{p}_{\min}=(-3.2,-4.8,-5.6)$. Only points whose voxel indices fall inside the valid grid are kept. 

\paragraph{Ground-truth occupancy voxels.}
The Occ-ScanNet ground truth is loaded from compressed \texttt{labels.npz} files. Each file contains semantic occupancy labels, camera visibility masks, LiDAR visibility masks, and, for raw Occ-ScanNet supervision, the raw semantic grid with its voxel origin and voxel size. We use the raw Occ-ScanNet ground truth branch during training and evaluation. The raw semantic grid has size $60 \times 60 \times 36$ at $0.08$ m together with a per-frame voxel origin and voxel size, and follows the native $[y,x,z]$ axis order. Voxels labelled $0$ are empty, labels $1,\ldots,11$ are the occupied semantic classes, and $255$ denotes unknown voxels, which are excluded from both training targets and evaluation. Raw semantic labels in the range $1,\ldots,11$ are treated as occupied semantic classes and converted to zero-based class IDs $0,\ldots,10$; free space and unknown voxels are ignored for the point-level semantic targets.

The raw Occ-ScanNet tensor is indexed in $(y,x,z)$ order. Therefore, when constructing 3D target points, a raw voxel coordinate $(i_y,i_x,i_z)$ is converted to Cartesian order as $(i_x,i_y,i_z)$. The raw voxel origin denotes the center of voxel $(0,0,0)$ rather than the minimum grid corner. Hence the 3D center of a raw occupied voxel is computed as
\[
\mathbf{p}_{\mathrm{gt}}
=
\mathbf{o}_{\mathrm{raw}}
+
(i_x,i_y,i_z)\odot\Delta_{\mathrm{raw}},
\]
where $\mathbf{o}_{\mathrm{raw}}$ is the raw voxel-origin vector and $\Delta_{\mathrm{raw}}$ is the raw voxel size. This convention avoids a half-voxel offset when matching predicted points to raw Occ-ScanNet supervision.

\paragraph{Prediction decoding and voxel aggregation.}
The decoder predicts normalized 3D refinement points inside the configured local range. We decode them back to metric coordinates in the occupancy frame. For raw Occ-ScanNet evaluation, predicted points are further transformed from the local occupancy frame to the world frame using
\[
T_{\mathrm{occ}\rightarrow\mathrm{world}}
=
T_{\mathrm{ego}\rightarrow\mathrm{world}}
T_{\mathrm{occ}\rightarrow\mathrm{ego}}.
\]
They are then mapped into the raw Occ-ScanNet grid using the raw voxel origin and voxel size. Since the raw grid origin is center-defined, the binning rule is
\[
\mathbf{i}_{\mathrm{raw}}
=
\left\lfloor
\frac{\mathbf{p}_{\mathrm{world}}-\mathbf{o}_{\mathrm{raw}}}
{\Delta_{\mathrm{raw}}}
+0.5
\right\rfloor .
\]
If multiple predicted points fall into the same voxel, we keep the top-$k$ points by their maximum semantic confidence, with $k=10$, and average their semantic scores to obtain the voxel-level prediction. During evaluation, points with semantic confidence below 0.4 are discarded, and points too far from their query center are removed using a center-distance threshold of 1.0 m. We additionally apply the configured sparse padding step to fill small holes before computing occupancy metrics.

\paragraph{Training target construction.}
For training, occupied raw voxels are converted to sparse 3D points and used as point-level targets. Predicted query points are matched to ground-truth occupied voxel centers using a bidirectional point matching objective. The point loss is a density-aware Chamfer-style loss implemented with Smooth-L1 distance, and semantic classification is supervised with focal loss. The final loss combines the last decoder layer and auxiliary decoder layers with an exponential layer decay of 0.9. A containment loss is additionally applied on decoder layers d4 and d5 to penalize predictions that fall into known empty regions and pulls them toward nearby occupied voxel boxes. Its weight is linearly increased from 0.1 to 0.3 between epochs 20 and 60.

\paragraph{Evaluation protocol.}
Evaluation uses the full 500-query budget. Predicted points are mapped from the local occupancy frame into the official Occ-ScanNet raw grid ($60 \times 60 \times 36$, voxel size $0.08$ m) using the per-frame voxel origin and voxel size, and compared against its raw semantic labels. Voxels with unknown labels ($255$) are excluded, which corresponds to the camera-visible region of the selected front camera; predictions falling outside the official grid are ignored. We report occupied IoU and mIoU over the 11 occupied semantic classes.

\begin{table}[h]
\centering
\small
\caption{Main implementation details for AdaOcc on Occ-ScanNet-mini.}
\label{tab:supp_impl_details}
\begin{tabular}{p{0.36\linewidth}p{0.52\linewidth}}
\toprule
\multicolumn{2}{c}{\textbf{Data and Input Configuration}} \\
\midrule
Dataset & Occ-ScanNet-mini \\
Training / validation samples & 4,639 / 2,007 \\
RGB observations $\mathcal{I}$ & Single-view RGB, current frame only \\
Geometric observations $\mathcal{G}$ & Depth-derived 3D points for benchmark experiments \\
Image resolution & $960 \times 720$ after resizing \\
Occupancy range $\Omega$ & $[-3.2,-4.8,-5.6,7.2,4.8,5.6]$ \\
Voxel size $\mathbf{v}$ & $[0.08,0.08,0.08]$ \\
Semantic classes & 11 occupied classes + free label \\
\midrule
\multicolumn{2}{c}{\textbf{Model Configuration}} \\
\midrule
Image encoder $E_{\mathrm{img}}$ & C-RADIOv3-B, last 4 blocks unfrozen \\
Image feature dimension & 512 \\
Geometric encoder $E_{\mathrm{geo}}$ & Sparse 3D convolutional encoder with multi-plane feature aggregation \\
Decoder layers $L$ & 6 \\
Maximum query budget $N_{\max}$ & 500 \\
Initial query budget $N_{\mathrm{init}}$ & 100 \\
Progressive query schedule $N_q(t)$ & Increase by 100 queries every 40 epochs until 500 \\
Queries at evaluation & Full query budget unless otherwise specified \\
Points per query refinement & $[1,2,4,8,16,32]$ across decoder layers \\
\midrule
\multicolumn{2}{c}{\textbf{Optimization}} \\
\midrule
Optimizer & AdamW \\
Base learning rate & $2\times10^{-4}$ \\
Weight decay & 0.01 \\
Batch size & 8 per GPU, 64 global \\
Training epochs & 200 \\
Learning-rate schedule & 500-iteration linear warmup + cosine decay to $2\times10^{-7}$ \\
Mixed precision & Enabled with static loss scale 512 \\
Gradient clipping & L2 norm clipping with max norm 35 \\
\midrule
\multicolumn{2}{c}{\textbf{Loss Configuration}} \\
\midrule
Semantic loss $\mathcal{L}_{\mathrm{sem}}$ 
& Focal loss with $\gamma_{\mathrm{focal}}=2.0$, $\alpha=0.25$, $\lambda_{\mathrm{sem}}=2.0$ \\
Point reconstruction $\mathcal{L}_{\mathrm{pts}}$ 
& DCD Loss with Smooth L1 regression, $\lambda_{\mathrm{pts}}=0.5$ \\
Decoder loss decay 
& $\gamma=0.9$ \\
Containment loss $\mathcal{L}_{\mathrm{con}}$ 
& Applied to layer 5 and 6, with $\lambda_{\mathrm{con}}$ ramped from 0.10 to 0.30 \\
\bottomrule
\end{tabular}
\end{table}

\subsection{Experimental Settings on Different Datasets}
We provide a summary of the experimental settings across datasets in Table~\ref{tab:dataset_comparison}. 
The compared settings cover different combinations of RGB input, LiDAR input, open-vocabulary support, depth map sources, and the number of available views.

\begin{table*}[t]
\centering
\caption{Experimental Settings on Different Datasets}
\vspace{0.8em}
\setlength{\tabcolsep}{5pt}
\renewcommand{\arraystretch}{1.25}
\resizebox{\textwidth}{!}{%
\begin{tabular}{
>{\centering\arraybackslash}m{3.0cm}
>{\centering\arraybackslash}m{2.2cm}
>{\centering\arraybackslash}m{2.2cm}
>{\centering\arraybackslash}m{2.8cm}
>{\centering\arraybackslash}m{2.6cm}
>{\centering\arraybackslash}m{2.2cm}
}
\hline
\textbf{Datasets} &
\textbf{RGB Input} &
\textbf{LiDAR Input} &
\textbf{Open Vocabulary} &
\textbf{Depth Map Type} &
\textbf{\#Views} \\
\hline

Occ-ScanNet & \checkmark &  &  & Pseudo & 1 \\
\hline
Occ3D-nuScenes (setting1) & \checkmark & \checkmark &  &  & 6 \\
\hline
Occ3D-nuScenes (setting2) & \checkmark &  &  & Pseudo & 6 \\
\hline
 TartanGround (setting1) & \checkmark & \checkmark & \checkmark &  & 1-4 \\
\hline
 TartanGround (setting2) & \checkmark &  & \checkmark & GT & 1-4 \\
\hline
\end{tabular}%
}
\label{tab:dataset_comparison}
\end{table*}

\subsection{Compute Resources}

The compute resources for the Occ-ScanNet-mini RADIO configuration are summarized in Table~\ref{tab:supp_compute_resources}. The model was trained on one internal Linux node with 8 NVIDIA H20 GPUs using PyTorch distributed data parallelism. Each GPU has approximately 97,871 MiB of memory. The 200-epoch training process took about 8.0 hours on 8 GPUs, corresponding to approximately 64 GPU-hours for training. Including the final validation pass, the total recorded run time was about 8.2 hours. The final 500-query training stage used about 11.0 GB memory per GPU, while validation used about 4.8 GB per GPU. Validation latency measured from the training log was approximately 0.34 s per sample per GPU, including data loading and metric computation.

\begin{table}[h]
\centering
\small
\caption{Compute resources for the Occ-ScanNet-mini AdaOcc run.}
\label{tab:supp_compute_resources}
\renewcommand{\arraystretch}{1.18}
\begin{tabular}{ll}
\toprule
Item & Value \\
\midrule
Hardware & 8 $\times$ NVIDIA H20 GPUs \\
GPU memory & 97,871 MiB per GPU \\
Training framework & PyTorch DDP with NCCL backend \\
CUDA / cuDNN & CUDA 12.1 / cuDNN 8.9.2 \\
PyTorch / TorchVision & 2.2.0 / 0.17.0 \\
Training time & 8.0 h on 8 GPUs \\
Total recorded run time & 8.2 h including final validation \\
Estimated training compute & 64 GPU-hours \\
Steady-state training memory & 11.0 GB per GPU at 500 queries \\
Validation memory & 4.8 GB per GPU \\
Validation latency & 0.34 s per sample per GPU \\
Checkpoint size & 3.4 GB per checkpoint \\
\bottomrule
\end{tabular}
\renewcommand{\arraystretch}{1.0}
\end{table}

\section{Limitations}
\label{sec:supp_limitations}

AdaOcc still has limited capability in modeling very small or thin objects. Since the method represents occupied regions with a finite set of semantic points and voxelizes them into a standard occupancy grid for evaluation, fine-grained structures may be under-represented when they occupy only a few voxels or are weakly observed from the input views. This limitation can be further affected by occlusion, noisy geometric inputs, or incomplete depth/LiDAR observations in cluttered indoor scenes. Beyond fine-grained geometry, AdaOcc estimates depth from RGB with an external model (Depth Anything v2), which adds inference latency and introduces failure modes on reflective or low-texture surfaces, thin structures, and occluded boundaries; replacing the estimate with ground-truth depth quantifies both the dependency and the remaining headroom (Appendix~\ref{sec:supp_depth}). Our evaluation is further limited to indoor benchmarks in the single-frame or direct-aggregation setting, so broader indoor evaluation and explicit temporal memory remain open directions. Future work will mitigate these issues through finer-grained query allocation, stronger local geometric refinement, small-object-aware supervision, and a more robust treatment of the geometric input.

\section{Broader Impacts}
\label{sec:supp_impacts}

AdaOcc aims to improve adaptive 3D scene perception for embodied agents. Its potential positive impacts include enabling robots and embodied systems to build more accurate spatial representations under varying sensing and computational conditions, which may benefit indoor navigation, assistive robotics, and other downstream embodied tasks. The budget-adaptive design may also make 3D perception more accessible on platforms with limited onboard computation.

At the same time, occupancy prediction systems may introduce risks when deployed in real-world environments. Incorrect occupancy or semantic predictions could affect downstream planning decisions, especially in safety-critical settings involving humans or fragile objects. In addition, 3D perception systems used in indoor environments may raise privacy concerns if raw sensor data are collected or stored. AdaOcc does not directly address these deployment-level issues. Practical use should therefore include task-specific safety checks, privacy-preserving data handling, and validation under the target sensing and operating conditions.

\section{Assets and Licenses}
\label{sec:supp_assets}

This work uses existing academic datasets, pretrained models, and baseline implementations for research evaluation. We cite the original sources for Occ-ScanNet, Occ-ScanNet-mini, TartanGround, Depth Anything v2, EfficientNet, RADIO/C-RADIO, OPUS, SplatSSC, and other compared methods in the main paper. These assets are used only for academic research and evaluation under their original licenses and terms of use. We do not redistribute the original datasets, pretrained checkpoints, or third-party code as part of this submission.


\end{document}